\documentclass[twoside,11pt]{article}

\usepackage{blindtext}

\usepackage[preprint]{jmlr2e}

\usepackage{amsmath}
\usepackage{amssymb}
\usepackage{mathtools}
\usepackage{enumitem}
\usepackage{graphicx}

\usepackage{pdflscape}

\usepackage{tikz}
\usetikzlibrary{arrows.meta, positioning}
\usepackage{longtable}
\usepackage{booktabs}
\usepackage{array}
\usetikzlibrary{calc}

\usepackage{microtype}
\usepackage{lastpage}
\jmlrheading{23}{2022}{1-\pageref{LastPage}}{1/21; Revised 5/22}{9/22}{21-0000}{Author One and Author Two}

\ShortHeadings{Unipolar Neurons for Reward-Based Continual Learning}{Leon}
\firstpageno{1}

\begin{document}

\title{Unipolar Neuron Networks Trained by Recirculation for Reward-Based Continual Learning}

\author{\name Florin Leon \email florin.leon@academic.tuiasi.ro \\
       \addr Faculty of Automatic Control and Computer Engineering \\
       ``Gheorghe Asachi'' Technical University of Iași \\
        Bd. Mangeron 27, 700050 Iași, Romania
     }


\maketitle

\begin{abstract}
Unipolar neurons are the predominant type of nerve cell in the octopus and in most other invertebrates. A single neurite (a fiber extending from the cell body) branches into a tree that carries both input and output synapses. The branches send graded local outputs to nearby cells, and one site on the main neurite sends a global output to distant targets. This paper proposes a network of such neurons and a learning rule for it. Each neuron is modeled as a tree of passive segments, whose activations are computed exactly as the steady state of a linear system. The network learns from reward alone, with a rule called recirculation. After each forward pass, a second pass sends requested changes of the outputs back through the same connections. Each synapse then adjusts its weight using only quantities available at its own location. Because the system of each neuron is symmetric, this update closely follows the gradient of the error in the predicted reward. Two mechanisms limit forgetting when tasks are learned in sequence: a task code inhibits a different subset of branches in each task, while consolidation lowers the learning rate of synapses that changed strongly in the recent past. The model is evaluated on a sequence of MNIST and Iris tasks. The network maintains high accuracy with low forgetting, whereas gradient-based training of the same network forgets earlier tasks almost completely.
\end{abstract}


\section{Introduction}

Most artificial neural networks model a neuron as a single point. The weighted inputs are summed and passed through a nonlinearity, and every target receives the same output. Biological neurons receive their inputs at different points of a dendritic tree. The position of a synapse affects how its input interacts with other inputs. In \emph{unipolar neurons}, the predominant type of nerve cell in the octopus and in most other invertebrates, a single neurite (a fiber extending from the cell body) branches into a tree. Input and output synapses lie side by side on the same branches. Action potentials start at one site of the main neurite and travel along the axon to distant targets. A unipolar neuron therefore has many graded local outputs, each of which mainly reflects the inputs to its own branch, and one global output that combines the contributions of all branches.

This paper applies the unipolar organization to continual learning, in which a network learns a sequence of tasks without access to the data of previous tasks. Networks trained by gradient descent tend to lose earlier tasks when they learn new ones, a problem known as catastrophic forgetting. Most methods that reduce forgetting add a separate component to the learning algorithm, such as a penalty on weight changes, a mask on the gradients or a buffer of stored examples. The network considered here learns without labels. At each step, it chooses an action and receives a reward that indicates whether the action was correct.

The proposed model is a network of unipolar neurons. Each neuron is a tree of passive segments, whose activations are the steady state of a linear system. The network learns with a two-pass rule called recirculation. The first pass computes the outputs and selects an action. The second pass sends requested changes of the outputs back through the same connections. Each synapse then changes its weight according to the quantities available at its own location. Because the system of each neuron is symmetric, this rule closely follows the gradient of the error in the predicted reward. Forgetting is limited by two mechanisms. A task code inhibits a different subset of branches at their base in each task. Consolidation lowers the learning rate of each synapse according to its recent changes. 
The inhibition of a branch has two effects. It reduces the contribution of the branch to the global output, so that each task uses partly distinct branches. It also attenuates the requested changes that reach the branch in the second pass, so that the weights learned in other tasks are largely preserved. Much of the protection thus follows from the structure of the neuron.

The main contributions of the paper are the following:

\begin{itemize}
\item A neuron model based on the organization of unipolar neurons, with graded local outputs, one global output and shunting or divisive inhibition, which scales down the effect of excitation at the same location. A single neuron of this type can compute the XOR function of two inputs, which a perceptron cannot;
\item Recirculation, a learning rule that uses only the reward and the quantities available at each synapse, and an analysis of its relation to the gradient;
\item Two mechanisms for continual learning, namely a task code that acts through dendritic inhibition and a consolidation rule that each synapse computes from its own history. Neither mechanism requires stored data or copies of earlier weights;
\item An experimental evaluation on a sequence of MNIST tasks and on the Iris dataset, including a comparison with gradient-based training of the same network, an analysis of the balance between learning new tasks and retaining earlier ones, and ablations of the main elements of the model.
\end{itemize}

The remainder of the paper is organized as follows. Section 2 reviews related work on unipolar neurons, biologically plausible learning rules and methods for continual learning. Section 3 describes the neuron model, the network structure, the learning rule, the mechanisms for continual learning, and some variants used for comparison. Section 4 presents the experimental results on MNIST and Iris, followed by a discussion of the findings. Section 5 concludes the paper and outlines directions for future work.

The implementation is available at: {\color{blue}https://github.com/florinleon/OctopusNeuron}.

\section{Related Work}
\label{sec:related-work}

\subsection{Unipolar Neurons}
\label{sec:related-unipolar}

To the best of our knowledge, unipolar neurons have not been used so far as units of machine learning models. In the neural network literature, the term \emph{unipolar} usually describes units whose activation lies between 0 and 1. Neuroscience studies have, however, built compartmental models of unipolar neurons in insects. A compartmental model divides a neuron into small connected segments and computes the voltage (activation) of each segment from its inputs and its neighbors, as the model of Section~\ref{sec:model-description} does. These studies examine whether a neuron behaves as one unit or as several partly independent units, and how inhibition and the location of the inputs affect its output.

\citet{meier2019extreme} studied a neuron of the fruit fly (Drosophila melanogaster) that receives visual input from the entire eye. The neuron
has a separate terminal for each small region of the visual field, and each terminal both receives input from that
region and sends output back to it. Recordings of activity showed that each terminal responds to stimuli in its own
region, independently of the neighboring terminals. A compartmental model indicated that the thin connections between
the terminals make this independence possible. A single neuron can therefore act as many local units, each with its
own inputs and outputs.

\citet{amin2020localized} studied a single inhibitory neuron in a brain region of the fruit fly that is involved in
learning odors. This neuron receives input from a large population of neurons and inhibits the same population in
return. It communicates through continuous changes of its voltage, without spikes. Activity evoked at one location
decayed with distance along its branches, and the inhibition was strongest near that location. A model of this decay
predicted that each neuron of the population is inhibited more by its own activity than by that of any other single
neuron. One inhibitory neuron thus acts as many local feedback loops.

\citet{liu2022connectomic} also built compartmental models of fruit fly neurons that receive olfactory information. Their
shapes and synapse positions came from the connectome, i.e., a map of synaptic connections reconstructed from electron
microscopy images. Models with many compartments matched the measured responses better than models that treat each
neuron as a point. Some of these neurons have two groups of branches joined by a long, thin cable. Because the cable
attenuates signals, each group weights its inputs with its own gain while sharing part of its activity with the other.
The authors suggested that this partial separation allows computations within a single neuron.

\citet{hafez2023cellular} built a compartmental model of an output neuron in the same learning region of the fruit
fly, again based on the connectome. Voltages differed strongly between the branches of the neuron, but every synapse
had nearly the same effect on its output. The neuron therefore sums its inputs regardless of where they arrive. The
authors argued that this property makes the readout of stored memories robust to the random wiring of the inputs.
Together with the previous study, this result indicates that the shape of the tree determines whether a neuron
combines its inputs locally or globally.

\citet{groschner2022biophysical} explained how a motion-sensitive neuron of the fruit fly computes a product of its
inputs, as classic models of motion detection require. Its excitatory and shunting inhibitory inputs arrive at
different locations of its dendrite. Shunting inhibition opens channels in the membrane with little change of its
voltage, so it divides the effect of excitation. When a moving stimulus removes this inhibition as excitation arrives,
the response exceeds the sum of the two separate effects. A passive model reproduced this multiplication without any
threshold. The preference for one direction of motion follows from the different arrival times of the spatially
separated inputs.

\citet{dewell2022contrast} studied a large neuron of the grasshopper that detects approaching objects and produces its
spikes at one site of its main branch. Inputs that signal darkening and inputs that signal brightening arrive on two
separate fields of its dendritic tree, one far from this site and one close to it. A biophysical model showed that the
inputs of the closer field are attenuated less on their way to the site, so fewer synapses are needed to produce
spikes. The location of a group of inputs within the tree is thus part of the computation of the neuron.

\citet{lappalainen2024connectome} built a network model of the motion-detecting part of the fruit fly visual system,
with connections taken from the connectome. Each neuron is a single compartment that produces graded outputs and no
spikes. The parameters of the neurons and the strengths of the synapses were trained by backpropagation, so that the
network estimates the motion in videos. The trained network predicted the responses of many types of neurons, and
these predictions agreed with experimental recordings. The study shows that connectivity taken from biology, combined
with optimization for a task, can reproduce the function of real neurons.

The present paper differs from these studies, which either analyze individual neurons without learning or train networks of single-compartment neurons. This paper proposes a model of such a neuron as the unit of a network that learns a sequence of tasks from reward.

\subsection{Continual Learning and Reinforcement Learning}
\label{sec:related-continual-rl}

\citet{khetarpal2022towards} reviewed continual reinforcement learning, which they framed as learning from reward in
environments that change over time. They classified the settings by the scope of the change, for example whether it
affects the rewards or the dynamics of the environment, and by the factors that drive it. They also surveyed the main
families of methods, the benchmarks and the evaluation metrics. The review discusses open problems, such as the balance
between retaining earlier knowledge and adapting to new conditions. It also relates continual learning in artificial
agents to findings on learning and memory in animals.

\citet{kaplanis2018continual} replaced the scalar weights of learning agents with a model of complex synapses. In this
model, each synapse has several hidden variables that interact and change on different timescales, so its effective
plasticity depends on its history. The authors applied the synaptic model to tabular and deep reinforcement learning
agents. It reduced forgetting when two simple tasks were learned in sequence, without any information about task
boundaries. Within a single task, it also reduced the need for an experience replay buffer, a memory of past transitions that is used for repeated training.

\citet{elsayed2024addressing} addressed catastrophic forgetting and the loss of plasticity, the gradual
decline in the ability of a network to learn. Their method, utility-based perturbed gradient descent, estimates the
utility of each weight as the increase in the loss that its removal would cause. Useful weights receive smaller updates,
which protects them, while noise added to less useful weights keeps them plastic. The method runs online without task
boundaries or replay. It performed comparably to or better than elastic weight
consolidation~\citep{kirkpatrick2017overcoming} and synaptic intelligence~\citep{zenke2017continual}. In a
policy-gradient agent, it also avoided the late decline of performance observed with the Adam optimizer.

\citet{beniwhiwhu2023lifelong} applied modulating masks to lifelong reinforcement learning. The network keeps its random
initial weights, and a separate mask over them is learned for each task. Because these shared parameters never change,
a new task cannot overwrite the earlier ones. The masks of previous tasks can be combined linearly to form the mask of a
new task, which speeds up learning and is beneficial when rewards are sparse. The method retained earlier tasks better than online
elastic weight consolidation and experience replay. It requires the identity of the task during training and testing,
and it stores one mask per task.

\citet{iyer2022avoiding} added active dendritic segments to the units of a deep network. Each segment receives a
context signal, such as a code of the current task, and the strongest dendritic response modulates the output of the
unit. A $k$-winner-take-all step, which keeps only the $k$ most active units of a layer, then selects a sparse
subnetwork for each context, and only its units are updated. The model was applied to multi-task reinforcement
learning, in which a robot learns several manipulation tasks with one network, and to continual supervised learning.
Different contexts selected overlapping but distinct subnetworks, which reduced interference between tasks.

The present paper differs from these approaches in two respects: the task code gates branches through shunting inhibition (inhibition that divides the effect of excitation at the same location instead of subtracting from it), which also attenuates the weight changes in the gated branches, and the network learns from reward with a local rule in which each synapse computes its own consolidation.

\section{Model Description}
\label{sec:model-description}

\subsection{Unipolar Neurons}
\label{sec:perceptron-unipolar}

\begin{figure}[t]
  \centering
  \includegraphics[width=\linewidth]{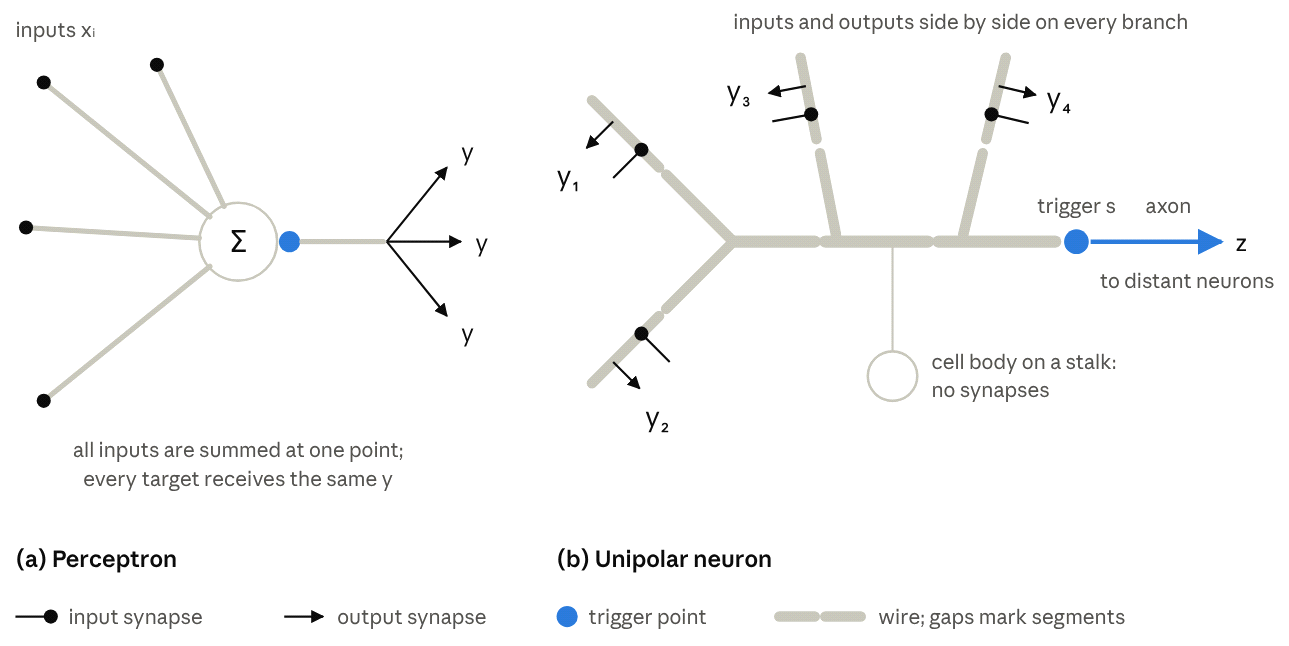}
  \caption{A perceptron (a) and a unipolar neuron (b). (a) All inputs $x_i$ are summed at one point, and every target
    receives the same output~$y$. (b) Input synapses and output synapses lie side by side along the branches; gaps in
    the branches mark the boundaries between segments. Each local output ($y_1$ to $y_4$) depends on the activation of
    its own segment and reaches nearby neurons. The trigger~$s$ sends the global output~$z$ along the axon to distant
    neurons. The cell body is attached to the trunk by a thin stalk, receives no synapses and is not part of the
    model.}
  \label{fig:perceptron-unipolar}
\end{figure}

A perceptron~\citep{rosenblatt1958perceptron} models a neuron as a single point. All inputs are summed at this point,
and one output is sent to every target (Figure~\ref{fig:perceptron-unipolar}a):
\begin{equation}
  y = \varphi\Bigl(\sum_i w_i x_i - \theta\Bigr).
  \label{eq:perceptron}
\end{equation}

Here $x_i$ are the inputs, $w_i$ their weights, $\theta$ a threshold, and $\varphi$ a nondecreasing nonlinearity such as
$\max(0,\cdot)$. Each weight sets how strongly its input contributes to the output, and every target receives the same
value~$y$.

The neuron used in this paper is a tree of connected segments (Figure~\ref{fig:perceptron-unipolar}b). Each segment has
an \emph{activation}, a number between 0 and 1 that depends on the synaptic inputs to the segment and on the activations
of its neighbors (Section~\ref{sec:neuron}). Input synapses sit on the segments. The branch segments also carry output
synapses, which send a graded \emph{local output}~$y_k$ that depends on the activation of their segment. Local outputs
reach only nearby neurons. One segment of the trunk, the \emph{trigger}~$s$, produces the \emph{global output}~$z$ of
the neuron. Only the global output reaches distant neurons and the readout of the network. A single neuron therefore has
many inputs, several local outputs and one global output.

Each synapse has a weight and a fixed location in the tree. The effect of an input is largest on the segment of its
synapse and decreases with every step along the tree. A local output therefore reflects mainly the inputs to its own
branch. The trigger receives a reduced contribution from every branch and combines the contributions of all branches.
All outputs of a neuron depend on the same weighted inputs, but each output reads them at a different location. In this
sense, a unipolar neuron behaves like a small layer of units that share their input weights. The weight of a synapse
sets the strength of its input, and its location sets how this strength is divided among the outputs. The locations are
drawn at random when the network is built and remain fixed; only the weights are learned.

The shape of the tree determines how independent the outputs are. When neighboring segments are weakly coupled, each
segment depends mainly on its own inputs, and the neuron behaves like several separate units. When they are strongly
coupled, all segments have similar activations, and the neuron behaves like a point neuron. Inputs can also be
inhibitory. An input that excites one segment and inhibits another can raise one output of the neuron and lower another.
With such wiring, a single neuron with three segments computes the exclusive-or (XOR) of two binary inputs, which no
perceptron can do (Section~\ref{sec:neuron}).

Unipolar neurons are the typical neuron type of the octopus and of most other
invertebrates~\citep{bullock1965structure,young1971anatomy}. A single neurite leaves the cell body and branches into a tree. Input and output synapses lie side by side on the same branches. In many invertebrate neurons, output synapses release transmitter in a graded way that depends on the local membrane potential, without action potentials~\citep{burrows1978graded}. Action potentials start at one site of the main process and travel along the axon to distant targets. The cell body is attached to the main neurite by a thin stalk and receives no synapses. 

The proposed model uses this organization and omits the cell body. It describes the neuron in mathematical terms suited to machine learning and does not reproduce the electrical dynamics of real neurons.

\subsection{The Neuron Model}
\label{sec:neuron}

\begin{figure}[t]
  \centering
  \includegraphics[width=\linewidth]{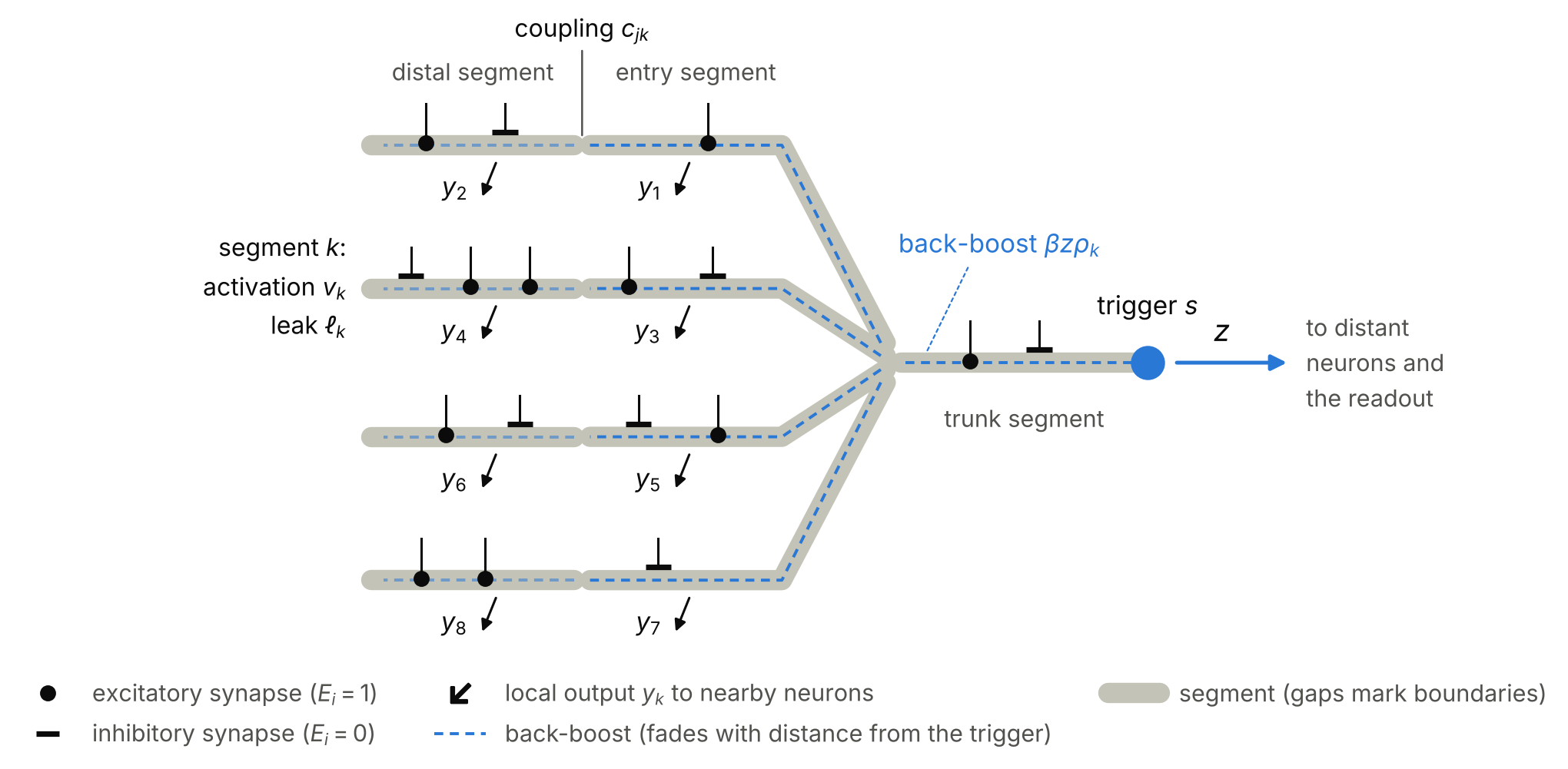}
  \caption{Structure of the default neuron. A single trunk segment contains the trigger~$s$, and four branches are
    attached to it, each made of an entry segment next to the trunk and a distal segment. Each segment $k$ has an
    activation~$v_k$ and a leak~$\ell_k$, and neighboring segments are linked by couplings~$c_{jk}$. Excitatory and
    inhibitory synapses can sit on any segment, including the trunk. Each of the eight branch segments sends a local
    output~$y_k$ to nearby neurons, and the trigger sends the global output~$z$ to distant neurons and the readout. The
    back-boost (dashed) adds $\beta z \rho_k$ to the activation of every segment~$k$ (Eq.~\ref{eq:boosted}); its
    lighter shade on the distal segments indicates that $\rho_k$ decreases with distance from the trigger.}
  \label{fig:neuron}
\end{figure}

A neuron is a tree of $K$ segments. One segment of the trunk, the trigger segment~$s$, produces the global output. In
the default neuron, a single trunk segment contains the trigger, and four branches of two segments each are attached to
it, so $K = 9$ (Figure~\ref{fig:neuron}). Each segment $k$ has an activation~$v_k$. Neighboring segments $j$ and $k$ are
linked by a coupling $c_{jk} > 0$, and each segment has a leak $\ell_k > 0$. Couplings and leaks describe the shape of
the neuron (e.g., leak 2 for the trunk, 0.5 for a branch segment, and coupling 2). They are fixed and not learned.

The inputs of the neuron arrive through synapses. Synapse $i$ is located on segment $k(i)$, receives an input $x_i \ge 0$ and
has a learned weight $w_i$ between 0 and an upper bound (e.g., 10). Its weighted input is $g_i = w_i x_i$. Each synapse
is either excitatory or inhibitory. The type is encoded by a fixed target value $E_i$, which is 1 for an excitatory
synapse and 0 for an inhibitory synapse. Every input source of a neuron (a pixel or an output of another neuron) forms
one excitatory and one inhibitory synapse on it, each on a segment drawn at random, including the trunk. A single input
can therefore excite one part of the tree and inhibit another. The model also allows several synapses of each type per
source, or synapses on branch segments only.

The activation of a segment is a weighted average of three kinds of terms: the target values of its synapses, the
activations of its neighbors, and 0. The weights of the average are the weighted inputs $g_i$, the couplings $c_{jk}$
and the leak $\ell_k$:
\begin{equation}
  v_k = \frac{\sum_{i \in k} g_i E_i + \sum_{j \sim k} c_{jk}\, v_j + \ell_k \cdot 0}
             {\sum_{i \in k} g_i + \sum_{j \sim k} c_{jk} + \ell_k}.
  \label{eq:activation}
\end{equation}

Here $i \in k$ runs over the synapses on segment $k$, and $j \sim k$ over its neighbors. Excitatory inputs move the
activation toward 1. Inhibitory inputs and the leak move it toward 0.

Eq.~\eqref{eq:activation} defines each activation through the activations of its neighbors. Multiplying it by its
denominator gives one linear equation per segment. Together, these equations form the system:
\begin{equation}
  \mathbf{M} \mathbf{v} = \mathbf{b}.
  \label{eq:system}
\end{equation}

Here $\mathbf{v}$ is the vector of all activations. The diagonal entries of $\mathbf{M}$ are the denominators of Eq.~\eqref{eq:activation},
$M_{kk} = \ell_k + \sum_{i \in k} g_i + \sum_{j \sim k} c_{jk}$. Two neighboring segments have $M_{jk} = -c_{jk}$, and
all other entries are zero. The right-hand side $b_k = \sum_{i \in k} g_i E_i$ is the excitatory weighted input of
segment $k$, because inhibitory synapses have $E_i = 0$. The matrix $\mathbf{M}$ is symmetric. In each row, the diagonal entry
exceeds the sum of the magnitudes of the other entries by at least the leak, so $\mathbf{M}$ is positive definite and the system
has a unique solution. All activations lie between 0 and 1: the largest activation is an average of numbers that are at
most 1 or at most itself, with a positive weight on 0, so it cannot exceed 1, and the same argument applied to the
smallest activation shows that it cannot be negative. Because the tree has no cycles, Gaussian elimination from the
leaves toward the trigger, followed by substitution back toward the leaves, solves the system exactly in a number of
operations proportional to $K$~\citep{hines1984efficient}. The model solves it for every input.

The activations are a nonlinear function of the inputs, because each weighted input appears both in the numerator and
in the denominator of Eq.~\eqref{eq:activation}, that is, both in $\mathbf{b}$ and in $\mathbf{M}$. Three properties follow from this
form.

The first two properties are easiest to see in a single segment without neighbors. Let this segment have leak $\ell$,
total excitatory weighted input $g_{\mathrm{exc}}$ (the sum of $g_i$ over its excitatory synapses) and total inhibitory
weighted input $g_{\mathrm{inh}}$. Eq.~\eqref{eq:activation} then reduces to:
\begin{equation}
  v = \frac{g_{\mathrm{exc}}}{g_{\mathrm{exc}} + g_{\mathrm{inh}} + \ell}.
  \label{eq:single}
\end{equation}

\begin{figure}[t]
  \centering
  \includegraphics[width=\linewidth]{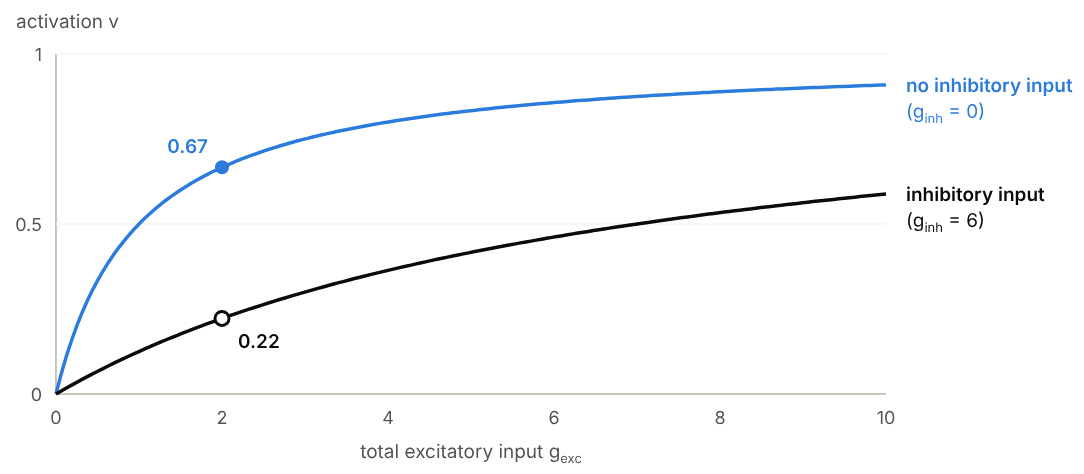}
  \caption{Activation of a single segment without neighbors as a function of its total excitatory weighted input
    $g_{\mathrm{exc}}$, for leak $\ell = 1$ (Eq.~\ref{eq:single}). Without inhibitory input (blue), the activation
    saturates toward 1. An inhibitory input $g_{\mathrm{inh}} = 6$ (black) scales the activation down. The markers show
    the activations for $g_{\mathrm{exc}} = 2$: 0.67 without inhibition and 0.22 with inhibition.}
  \label{fig:segment}
\end{figure}

First, the activation saturates. For $\ell = 1$, total excitatory inputs of 1, 2 and 10 give activations of 0.50, 0.67
and 0.91 (Figure~\ref{fig:segment}). The activation approaches 1 but never exceeds it. Second, inhibition acts
divisively. The inhibitory input appears only in the denominator, and Eq.~\eqref{eq:single} can be written as:
\begin{equation}
  v = \frac{g_{\mathrm{exc}}}{g_{\mathrm{exc}} + \ell} \cdot
      \frac{g_{\mathrm{exc}} + \ell}{g_{\mathrm{exc}} + g_{\mathrm{inh}} + \ell}.
  \label{eq:divisive}
\end{equation}

The first factor is the activation without inhibition. The second factor lies between 0 and 1 and scales it down. For
$\ell = 1$ and $g_{\mathrm{exc}} = 2$, an inhibitory input of 6 multiplies the activation by $3/9$, from 0.67 to 0.22.
Without excitatory input, the same inhibitory input has no effect, because the activation is already 0. Excitation and
inhibition at the same location therefore interact multiplicatively.

The third property concerns segments without synapses. By Eq.~\eqref{eq:activation}, the activation of such a segment
is a weighted average of the activations of its neighbors and 0, so it is smaller than the activation of its most active
neighbor. Activation therefore decreases with every step away from the synapses that produce it, and it decreases faster
when the leaks are large compared with the couplings. The trigger receives an attenuated contribution from every branch
and combines the contributions of all branches. The location of inhibition matters in the same way. Inhibition on the
trunk enters only the equation of the trigger segment, so it lowers mainly the trigger activation and changes the branch
activations little. Inhibition on a branch lowers the activations of that branch and, with them, its contribution to the
trigger.

Each branch segment $k$ produces a local output, a rectified linear function (ReLU) of its activation:
\begin{equation}
  y_k = \max\left(0,\ v_k - t_k\right).
  \label{eq:local}
\end{equation}

Here $t_k$ is the threshold of output $k$. In the default neuron, all eight branch segments have local outputs; they can
also be restricted to the branch tips. The trigger segment produces the global output:
\begin{equation}
  z = \sigma\!\left(\frac{v_s - \theta}{T}\right),
  \label{eq:global}
\end{equation}
where $\sigma$ is the logistic function, $\theta$ is the threshold of the neuron, and $T$ is a temperature that sets
the slope.
The global output lies between 0 and 1 and equals 0.5 when $v_s = \theta$. It is a real number and a deterministic
function of the input; no spikes are sampled. The thresholds and the temperature are set before training by calibration
(Section~\ref{sec:network}) and then remain fixed.

\begin{figure}[t]
  \centering
  \includegraphics[width=\linewidth]{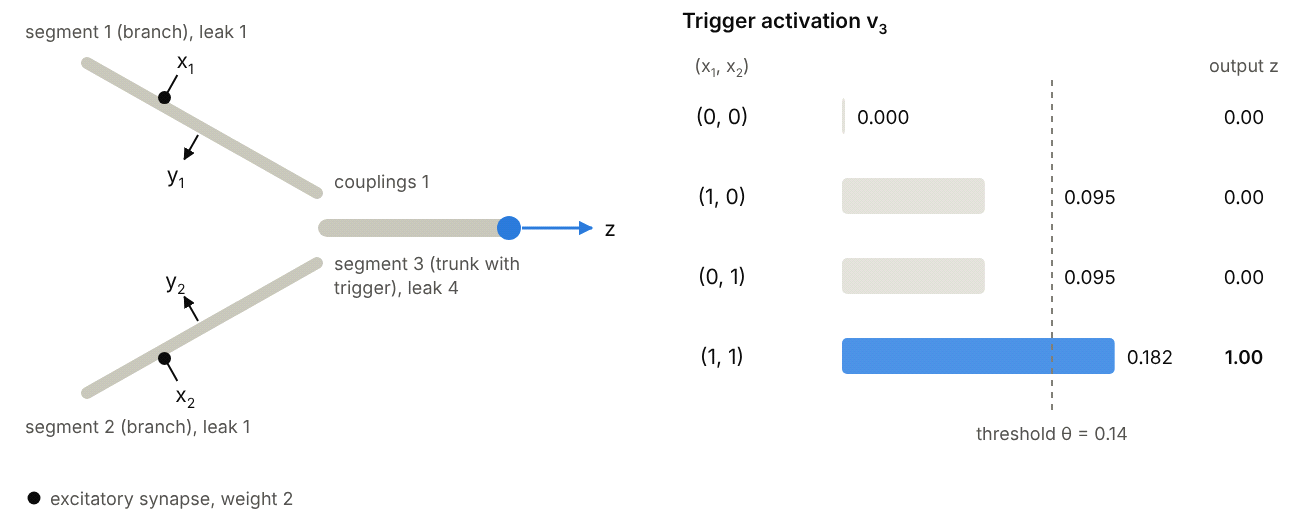}
  \caption{Example 1: AND. Each input of the three-segment neuron has an excitatory synapse with weight 2 on its own
    branch. The bars show the trigger activation $v_3$ for each input combination. The right column shows the global
    output $z$ for $\theta = 0.14$ and $T = 0.002$.}
  \label{fig:example-and}
\end{figure}

\paragraph{Example 1: AND.}
Both this example and the next use a neuron with 3 segments (Figure~\ref{fig:example-and}). Segments 1 and 2 are branches with leak 1. Segment 3 is the trunk with leak 4 and contains the trigger. Each branch is linked to the trunk with coupling 1. The neuron has two binary inputs, $x_1$ and $x_2$. The examples omit the back-boost defined below, so that all values can be computed more easily. In Example 1, $x_1$ has an excitatory synapse on segment 1 and $x_2$ has one on segment 2.
Both synapses have weight 2, so an active input contributes a weighted input of 2.

The trigger segment has no synapses. By Eq.~\eqref{eq:activation}, its activation is the weighted average of the two
branch activations and 0, with weights 1, 1 and 4:
\begin{equation*}
  v_3 = \frac{1 \cdot v_1 + 1 \cdot v_2}{1 + 1 + 4} = \frac{v_1 + v_2}{6}.
\end{equation*}

The trigger activation is therefore one sixth of the sum of the two branch activations.

Table~\ref{tab:example-and} lists all activations and outputs. When only $x_1$ is active, the activation of segment 1
is 0.524. The coupling passes activation on to the trunk, where it is 0.095, and from there to segment 2, where it is
0.048. When both inputs are active, each branch has activation 0.545, and the trigger activation is 0.182.

The thresholds determine which function each output computes. With output thresholds of 0.3, $y_1$ is positive exactly
when $x_1$ is active, and $y_2$ exactly when $x_2$ is active. With $\theta = 0.14$ and $T = 0.002$, the global output
$z$ is 1.00 when both inputs are active and 0.00 otherwise. The global output therefore computes $x_1$ AND $x_2$, while
each local output passes on one input.

\begin{table}[t]
  \centering
  \caption{Activations and outputs in Example 1 (output thresholds 0.3, $\theta = 0.14$, $T = 0.002$).}
  \label{tab:example-and}
  \begin{tabular}{ccccccc}
    \hline
    Inputs $(x_1, x_2)$ & $v_1$ & $v_2$ & $v_3$ & $y_1$ & $y_2$ & $z$ \\
    \hline
    $(0, 0)$ & 0     & 0     & 0     & 0     & 0     & 0.00 \\
    $(1, 0)$ & 0.524 & 0.048 & 0.095 & 0.224 & 0     & 0.00 \\
    $(0, 1)$ & 0.048 & 0.524 & 0.095 & 0     & 0.224 & 0.00 \\
    $(1, 1)$ & 0.545 & 0.545 & 0.182 & 0.245 & 0.245 & 1.00 \\
    \hline
  \end{tabular}
\end{table}

\paragraph{Example 2: XOR.}
The XOR of two binary inputs is 1 when exactly one of them is active. A neuron with only excitatory synapses cannot compute it, because adding an excitatory input increases the activation of every segment, so the trigger activation is always highest when both inputs are active.

\begin{figure}[t]
  \centering
  \includegraphics[width=\linewidth]{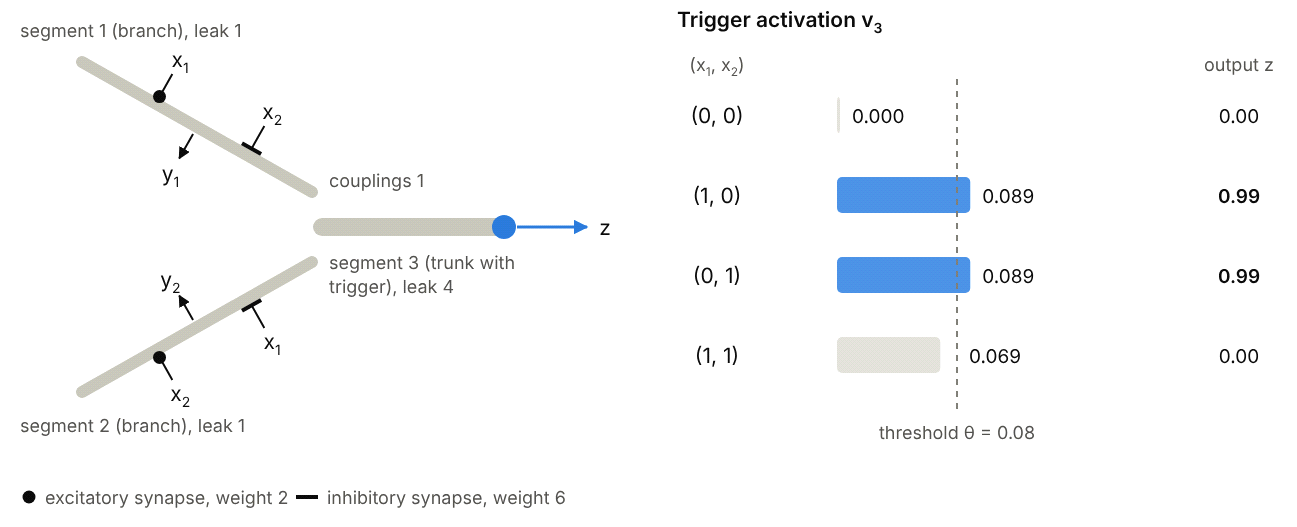}
  \caption{Example 2: XOR. Each input of the three-segment neuron has an excitatory synapse with weight 2 on its own
    branch and an inhibitory synapse with weight 6 on the other branch. The bars show the trigger activation $v_3$ for
    each input combination. The right column shows the global output $z$ for $\theta = 0.08$ and $T = 0.002$.}
  \label{fig:example-xor}
\end{figure}

Example 2 changes the wiring in one respect (Figure~\ref{fig:example-xor}): each input also has an inhibitory synapse with weight 6 on the other branch. Input $x_1$ thus excites segment 1 and inhibits segment 2, and $x_2$ excites segment 2 and inhibits segment 1. Because every input forms one excitatory and one inhibitory synapse on randomly chosen segments of each neuron that receives it, this kind of wiring arises in the network.

With one active input, for example $x_1 = 1$ and $x_2 = 0$, segment 1 receives an excitatory input of 2 and no
inhibitory input. By Eq.~\eqref{eq:activation}, its activation and the trigger activation are:
\begin{align*}
  v_1 &= \frac{2 \cdot 1 + 1 \cdot v_3}{2 + 1 + 1} = 0.522 \\
  v_3 &= \frac{v_1 + v_2}{6} = 0.089.
\end{align*}

Segment 2 receives an inhibitory input of 6 but no excitatory input. Its activation comes only from the trunk and is
small (0.011), so the inhibition has little effect. The trigger activation, 0.089, is close to its value in Example 1
(0.095).

With both inputs active, each branch receives an excitatory input of 2 and an inhibitory input of 6. The inhibitory
input adds 6 to the denominator of Eq.~\eqref{eq:activation}:
\begin{align*}
  v_1 &= \frac{2 \cdot 1 + 1 \cdot v_3}{2 + 6 + 1 + 1} = 0.207 \\
  v_3 &= \frac{0.207 + 0.207}{6} = 0.069.
\end{align*}

Each branch activation falls from 0.522 to 0.207. The sum of the two branch activations, 0.414, is now smaller than the
sum with a single active input, 0.533. The trigger activation therefore decreases from 0.089 to 0.069.

The trigger activations for the input combinations $(0, 0)$, $(1, 0)$, $(0, 1)$ and $(1, 1)$ are 0, 0.089, 0.089 and
0.069. Any threshold between 0.069 and 0.089 separates the two combinations with exactly one active input from the other
two. With $\theta = 0.08$ and $T = 0.002$, the global output $z$ is 0.00, 0.99, 0.99 and 0.00, which is the XOR of the
inputs (Table~\ref{tab:example-xor}).

The local outputs compute two further functions. With output thresholds of 0.3, $y_1$ is positive only for the
combination $(1, 0)$. When both inputs are active, the inhibition from $x_2$ keeps the activation of segment 1 below the
threshold. The local output $y_1$ therefore computes $x_1$ AND NOT $x_2$, and $y_2$ computes $x_2$ AND NOT $x_1$. A
single neuron with four synapses thus computes three different functions of its two inputs.

\begin{table}[t]
  \centering
  \caption{Activations and outputs in Example 2 (output thresholds 0.3, $\theta = 0.08$, $T = 0.002$).}
  \label{tab:example-xor}
  \begin{tabular}{ccccccc}
    \hline
    Inputs $(x_1, x_2)$ & $v_1$ & $v_2$ & $v_3$ & $y_1$ & $y_2$ & $z$ \\
    \hline
    $(0, 0)$ & 0     & 0     & 0     & 0     & 0     & 0.00 \\
    $(1, 0)$ & 0.522 & 0.011 & 0.089 & 0.222 & 0     & 0.99 \\
    $(0, 1)$ & 0.011 & 0.522 & 0.089 & 0     & 0.222 & 0.99 \\
    $(1, 1)$ & 0.207 & 0.207 & 0.069 & 0     & 0     & 0.00 \\
    \hline
  \end{tabular}
\end{table}

In the full model, the global output also acts on the branches, a process called \textit{back-boost}. This effect is described with the spread vector:
\begin{equation}
  \boldsymbol{\rho} = \mathbf{M}^{-1} \mathbf{e}_s,
  \label{eq:spread}
\end{equation}
where $\mathbf{e}_s$ is the unit vector of the trigger segment. If the right-hand side of the trigger segment in
Eq.~\eqref{eq:system} grows by a small amount $\epsilon$, the solution changes by $\epsilon \boldsymbol{\rho}$. The entry $\rho_k$ is
therefore the change of the activation of segment $k$ per unit of additive input at the trigger. Because $\mathbf{M}$ is
symmetric, $\rho_k = (\mathbf{M}^{-1})_{ks} = (\mathbf{M}^{-1})_{sk}$ is also the change of the trigger activation per unit of additive
input at segment $k$. The spread vector thus measures how strongly each segment and the trigger influence each other. It
depends on the input, because the weighted inputs enter $\mathbf{M}$. A strongly inhibited segment has a large diagonal entry,
and both this segment and the segments beyond it have small entries $\rho_k$.

The back-boost adds to every segment the change $\beta z \rho_k$ that an additive input $\beta z$ at the trigger would
produce, where $\beta$ is the back-boost strength (e.g., 2). The local outputs read the boosted activations:
\begin{equation}
  y_k = \max\left(0,\ v_k + \beta z \rho_k - t_k\right).
  \label{eq:boosted}
\end{equation}

In the full model, Eq.~\eqref{eq:boosted} replaces Eq.~\eqref{eq:local}. The global output is
computed from the activation of the trigger before the boost, so the boost does not change $z$. Through the back-boost,
each local output combines the inputs to its own branch with the global output of the neuron. When the global output is
high, all local outputs rise, most in branches that are strongly connected to the trigger and least in inhibited
branches. The spread vector appears again in learning, because requested changes of the global output enter at the
trigger and reach the segments in proportion to $\boldsymbol{\rho}$ (Section~\ref{sec:learning}).

Eq.~\eqref{eq:activation} is the steady state of a passive compartmental model of a
neuron~\citep{koch1999biophysics}. Each segment is a compartment, and $v_k$ is its membrane potential relative to rest,
scaled so that the reversal potential of excitatory synapses is 1. The weighted inputs $g_i$ are synaptic conductances,
the target values $E_i$ are reversal potentials, and the couplings and leaks are the axial and leak conductances.
Eq.~\eqref{eq:activation} then states that the currents into each compartment sum to zero. Inhibition whose
reversal potential equals the resting potential divides the effect of excitation, as in Eq.~\eqref{eq:single}, and is
known as shunting inhibition. The weights are nonnegative because a conductance cannot be negative. The global
output can be read as the firing probability of the trigger, and the back-boost as a steady-state counterpart of the
action potential, which spreads from the trigger back into the branches.

\subsection{The Network Model}
\label{sec:network}

The network is defined independently of the task. It needs only the number of inputs and their positions, the number of tasks and the number of actions. Neurons are arranged in layers. Each layer places groups of neurons on a
regular grid of positions in the unit square (e.g., 8 neurons at each of $7 \times 7$ positions).  The local outputs of a neuron take the position of their neuron.

A neuron receives one or more groups of inputs, each drawn from one population of sources: the pixels in an image, the local
outputs of an earlier layer, or the global outputs of an earlier layer. From each group, the neuron samples a fixed
number of sources once, when the network is built. Inputs and local outputs are sampled within a square window around
the position of the neuron, and global outputs are sampled from anywhere. Each sampled source then forms one excitatory
and one inhibitory synapse on the neuron (Section~\ref{sec:neuron}). This wiring gives the network a receptive-field
structure similar to that of a convolutional network, but without weight sharing: every neuron learns its own weights.

\begin{figure}[t]
  \centering
  \includegraphics[width=\linewidth]{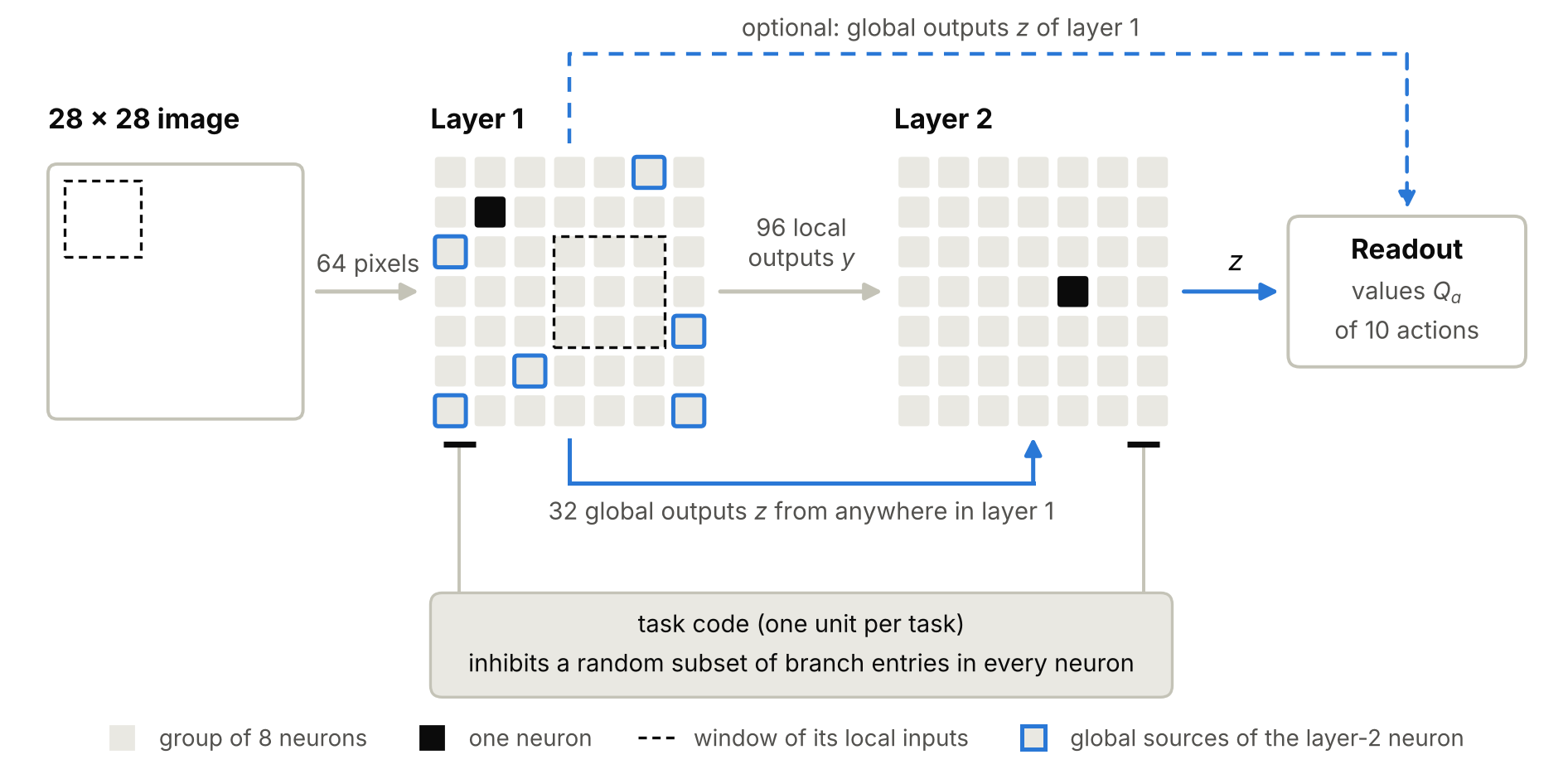}
  \caption{Default network for MNIST. Each layer is a grid of $7 \times 7$ positions with 8 neurons at each. A neuron
    of layer 1 samples 64 pixels from a window around its position. A neuron of layer 2 samples 96 local outputs of
    layer-1 neurons from a window around its position, and 32 global outputs of layer-1 neurons from anywhere. The
    global outputs of layer 2, and optionally those of layer 1 (dashed), reach the readout, which predicts the reward of
    each of the 10 actions. The task code inhibits a random subset of branch entries in every neuron of both layers.}
  \label{fig:network}
\end{figure}

The default network has two layers (Figure~\ref{fig:network}). If the input is an image, like in MNIST, a neuron of the first layer samples pixels near its
position (e.g., 64 pixels from a window of about $8 \times 8$ pixels). A neuron of the second layer samples local
outputs of nearby first-layer neurons (e.g., 96) and global outputs of first-layer neurons from anywhere (e.g., 32).
Local outputs therefore pass detailed, spatially restricted information forward, and global outputs provide a compact
summary of the whole image. 
Networks with more layers or other combinations of sources are built in the same way, and the learning rule applies to them without modification (Section~\ref{sec:learning}).

When the identity of the current task is available, it reaches the network as a one-hot task code, a vector with one
unit per task in which only the unit of the current task is active. This code is
given during both training and testing. Each code unit acts on every neuron through fixed inhibitory synapses at
the entry segments of its branches. For every combination of neuron, code unit and branch, such a synapse exists with
a fixed probability (e.g., 0.5), drawn once when the network is built. Its weight is fixed (e.g., 6), not
learned. An active code unit therefore adds a large inhibitory weighted input to the entry segments it reaches, in the same way as any inhibitory synapse in Eq.~\eqref{eq:activation}.

Strong inhibition at the entry segment separates a branch from the trunk. Inputs to the gated branch then contribute
little to the trigger activation and thus to the global output. Under a given task code, about half of the branches of
each neuron are gated. Different codes gate different branches, so the same input drives partly distinct branches and
neurons under different tasks. The scheme resembles context-dependent gating of hidden units~\citep{masse2018alleviating,leon2026biologically}, applied here to the branches within each neuron. Optionally, the code can also
inhibit the trigger segment of a random subset of neurons, which silences their global output while their branches
continue to work. Section~\ref{sec:continual} shows how gating also protects the weights of gated branches during learning. 

In biophysical terms, each gate is a shunting inhibitory synapse at the base of a branch, which isolates the branch from the trunk.

The readout is linear in the global outputs of the last layer. It assigns a value to each action $a$:
\begin{equation}
  Q_a = \sum_c u_{ac}\, z_c.
  \label{eq:readout}
\end{equation}

The sum runs over the neurons $c$ whose global outputs reach the readout, and $u_{ac}$ are the readout weights, which start at zero. Optionally, the global outputs of all layers reach the readout. Local outputs do not. For one-step episodes, as in the experiments, $Q_a$ predicts the reward of action $a$ for the current input. The readout has no bias term; a bias is shared by all inputs, so rewarding the classes of a new task would raise their values for the inputs of earlier tasks as well.

During training, the network chooses its action from the policy:
\begin{equation}
  \pi_a = (1 - \varepsilon)\, \frac{\exp(Q_a / \tau)}{\sum_{b} \exp(Q_b / \tau)} + \frac{\varepsilon}{A},
  \label{eq:policy}
\end{equation}
a softmax over the values with temperature $\tau$ (e.g., 0.1), mixed with a small share $\varepsilon$ (e.g., 0.1) of uniformly random actions. Here $A$ is the number of actions (e.g., 10 for MNIST). Because the forward pass is deterministic, the action choice is the only source of exploration. The uniform share keeps every action in use, including actions that a policy shaped by earlier tasks would almost never choose, such as the classes of a new task. At test time, the network chooses the action with the largest value.

\subsection{Learning by Recirculation}
\label{sec:learning}

\begin{figure}[t]
  \centering
  \includegraphics[width=\linewidth]{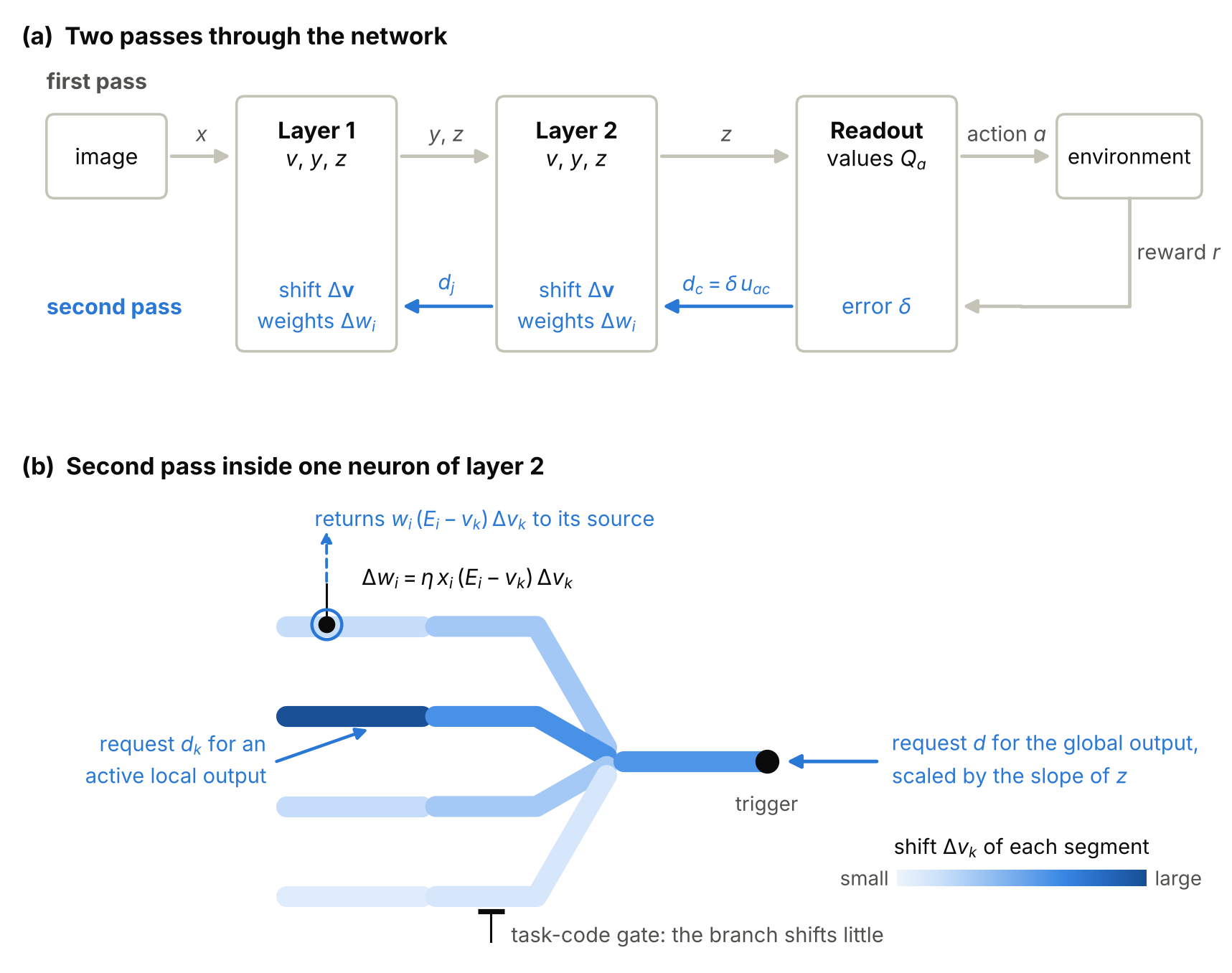}
  \caption{One learning step. In the first pass, the image passes through both layers to the readout, which predicts
    the reward of every action; the network chooses an action and receives a reward. In the second pass, the readout
    error $\delta$ of the chosen action returns as requested changes of the global outputs. Each neuron shifts its
    segment activations by $\Delta \mathbf{v}$ in response, each synapse changes in proportion to its input, its
    driving term $E_i - v_k$ and the shift of its segment, and the synapses pass requests back to the outputs of the
    earlier layer.}
  \label{fig:passes}
\end{figure}

The network learns online and from reward alone, with two passes through the network at every step
(Figure~\ref{fig:passes}). The first pass is the forward pass of Sections~\ref{sec:neuron} and~\ref{sec:network}: the network computes all activations and outputs, chooses an action and receives a reward. The second pass starts at the readout, which compares its prediction with the outcome and signals each neuron that it reads to change its global output in proportion to the error. These requested changes travel back through the same connections that carried activity forward, from the last layer to the first. Each neuron responds to the requests it receives with a modification of its segment activations. Each synapse then changes according to its input, its type and the modification of its own segment. The rule and the signals it uses are the same in every layer, so additional layers do not require any adjustments.

The name of the method refers to the recirculation procedure of ~\citet{hinton1988learning}, in which each weight changes in proportion to its presynaptic activity and to the difference between the postsynaptic activities in two passes. \citet{oreilly1996biologically} generalized this procedure to recurrent networks that learn arbitrary input--output mappings from local activation differences. The second pass is also related to the second phase of equilibrium propagation~\citep{scellier2017equilibrium}, in which the outputs are nudged toward a better value and the resulting change of the steady state determines the weight updates.

After the network chooses an action $a$, the environment returns a reward $r$. The readout compares its prediction with the outcome:
\begin{equation}
  \delta = r + \gamma V' - Q_a.
  \label{eq:error}
\end{equation}

Here $V' = \sum_b \pi'_b Q'_b$ is the value of the next input under the policy of Eq.~\eqref{eq:policy}, $\gamma$ is the discount factor (e.g., 0.9), and $V'$ is 0 when the episode ends. As in temporal-difference learning, the term
$r + \gamma V'$ is treated as a fixed target. In the experiments, every episode has a single step, so the error is
simply $\delta = r - Q_a$, the difference between the reward and its prediction. Only the chosen action has an error.

To move $Q_a$ toward the outcome, the readout demands each neuron $c$ that it reads to change its global output by:
\begin{equation}
  d_c = \delta\, u_{ac}.
  \label{eq:request-readout}
\end{equation}

When the prediction is too low ($\delta > 0$), neurons with a positive weight to the chosen action are demanded to become more active, and neurons with a negative weight less active. The readout thus sends the error back through each of its connections, with a strength equal to the forward weight of that connection. When the global outputs of all layers reach the readout, the neurons of earlier layers receive such requests directly, in addition to the requests that come from later layers.

A neuron can receive a request $d$ for its global output, from the readout or from neurons of later layers, and requests $d_k$ for its local outputs, from nearby neurons of the next layer. It turns all requests into additive terms
$\Delta \mathbf{b}$ on the right-hand side of its activation system~\eqref{eq:system}. At each output segment $k$:
\begin{equation}
  \Delta b_k =
  \begin{cases}
    d_k & \text{if } y_k > 0, \\
    0   & \text{otherwise}.
  \end{cases}
  \label{eq:db-local}
\end{equation}

An active local output follows the activation of its segment one to one (Eq.~\ref{eq:boosted}), so the request enters
at that segment unchanged. An inactive output lies on the flat part of its rectifier function and passes nothing back. At the trigger segment:
\begin{equation}
  \Delta b_s = \frac{z (1 - z)}{T} \Bigl( d + \beta \sum_k \rho_k\, \Delta b_k \Bigr).
  \label{eq:db-trigger}
\end{equation}

The factor $z(1 - z)/T$ is the slope of the global output in Eq.~\eqref{eq:global}, that is, how much $z$ changes per unit of trigger activation. The sum accounts for the back-boost. Every local output rises by $\beta \rho_k$ per unit of global output (Eq.~\ref{eq:boosted}), so requests for the local outputs also signal the global output to change.

The neuron responds with the change:
\begin{equation}
  \Delta \mathbf{v} = \mathbf{M}^{-1} \Delta \mathbf{b},
  \label{eq:shift}
\end{equation}
which is the change of the solution of Eq.~\eqref{eq:system} when $\Delta \mathbf{b}$ is added to its right-hand side. The weighted inputs do not change between the passes, so $\mathbf{M}$ is the matrix of the forward pass and the shift is exactly linear in the requests. The overall size of the requests therefore only scales all weight changes and is absorbed into the learning rate. The part of $\Delta \mathbf{b}$ that enters at the trigger spreads into the tree along the spread vector $\boldsymbol{\rho}$ of Eq.~\eqref{eq:spread}, and each part that enters at an output segment spreads from that segment. Strongly inhibited branches have small entries in $\boldsymbol{\rho}$, so they shift little in response to requests for the global output (Section~\ref{sec:continual}).

Each synapse compares the activation of its own segment in the two passes. Synapse $i$ on segment $k = k(i)$ changes its weight by:
\begin{equation}
  \Delta w_i = \eta\, x_i \left( E_i - v_k \right) \Delta v_k,
  \label{eq:weight}
\end{equation}
where $\eta$ is the learning rate. The three factors have a direct interpretation. The input $x_i$ restricts
learning to synapses that were active. The driving term $E_i - v_k$ is the direction in which a stronger synapse would move its segment: upward for an excitatory synapse, since $v_k < 1$, and downward for an inhibitory synapse, since $v_k > 0$. The shift $\Delta v_k$ is the direction in which the segment is asked to move. A synapse whose driving term agrees with the requested shift is strengthened, and a synapse whose driving term opposes it is weakened. When a segment is asked to rise, its active excitatory synapses grow stronger and its active inhibitory synapses grow weaker; when it is asked to fall, the reverse happens. All three factors are available at the synapse itself. Weights are kept between 0 and their upper bound, and all updates are averaged over a batch of episodes that run in parallel (e.g., 32). 
Section~\ref{sec:continual} adds a factor that slows down synapses that have changed strongly in the past.

The synapses also pass requests back to their sources. Raising the activity of source $j$ by a small amount raises the weighted input of each of its synapses in proportion to the weight $w_i$, and each of these synapses moves its segment in the direction $E_i - v_k$. The change requested of source $j$ is therefore:
\begin{equation}
  d_j = \sum_{i \in S_j} w_i \left( E_i - v_{k(i)} \right) \Delta v_{k(i)},
  \label{eq:request-source}
\end{equation}
where $S_j$ is the set of synapses of source $j$ on all the neurons that sample it. Each contact multiplies the factor
$(E_i - v_k) \Delta v_k$ that drives its own weight change by its forward weight instead of its input, so its return
strength equals its forward weight. When the source is a local output of an earlier neuron, $d_j$ reaches that neuron as a request for this local output. When the source is a global output, $d_j$ reaches it as a request for the global output. The earlier layer then responds in the same way, by Eqs.~\eqref{eq:db-local}--\eqref{eq:request-source}, so the requests travel back layer by layer, until the first layer that sends nothing back.

The second pass computes the gradient of the squared readout error $L = \delta^2 / 2$, with one omission. Consider a synapse on segment $k$. Raising its weighted input $g_i$ by a small amount $\epsilon$ adds $\epsilon$ to the diagonal entry $M_{kk}$ and $\epsilon E_i$ to $b_k$, so to first order the activations change by
$\epsilon (E_i - v_k) \mathbf{M}^{-1} \mathbf{e}_k$. The derivative of $L$ with respect to $g_i$ is therefore:
\begin{equation}
(E_i - v_k) \bigl(\mathbf{M}^{-1} \partial L / \partial \mathbf{v}\bigr)_k, 
\end{equation}
where $\partial L / \partial \mathbf{v}$
contains the direct derivatives of $L$ with respect to the activations through the outputs of the neuron. By the chain rule, these direct derivatives are exactly $-\Delta \mathbf{b}$ of Eqs.~\eqref{eq:db-local} and~\eqref{eq:db-trigger}, because each request is the negative derivative of $L$ with respect to the output it addresses. Since $\mathbf{M}$ is symmetric:
\begin{equation}
\mathbf{M}^{-1} \partial L / \partial \mathbf{v} = -\Delta \mathbf{v}, 
\end{equation}
and:
\begin{equation}
  -\frac{\partial L}{\partial g_i} = \left( E_i - v_k \right) \Delta v_k.
  \label{eq:gradient}
\end{equation}

With $g_i = w_i x_i$, Eq.~\eqref{eq:weight} is a step of gradient descent on $L$, and Eq.~\eqref{eq:request-source} gives the negative derivative of $L$ with respect to each source. The symmetry of $\mathbf{M}$ is what allows the error to spread through a neuron by the same rule as activity, so that no separate backward computation is needed inside the neuron.

The omission concerns the back-boost. The weighted inputs also change the spread vector $\boldsymbol{\rho}$, and with
it the boost $\beta z \rho_k$ that each local output receives. The second pass leaves this dependence out. It has no
effect in the last layer, whose local outputs receive no requests, so there the update follows the gradient exactly. In
earlier layers, the update approximates the gradient. Apart from this term, the update departs from gradient descent
only through the consolidation factor of Section~\ref{sec:continual} and the bounds on the weights.
Section~\ref{sec:variants} describes training the same network with the exact gradient.

After the second pass, the readout moves the prediction of the chosen action toward the outcome by a normalized
least-mean-squares step:
\begin{equation}
  \Delta u_{ac} = \alpha\, \frac{\delta\, z_c}{1 + \sum_{c'} z_{c'}^2},
  \label{eq:nlms}
\end{equation}
where $\alpha$ is the step size (e.g., 0.5). The weights of the other actions do not change. The update follows the
second pass because the requests of Eq.~\eqref{eq:request-readout} must use the weights that made the prediction. The normalization makes the change of $Q_a$ almost independent of how many neurons are active: after the step, $Q_a$ has moved toward the outcome by about the fraction $\alpha$ of the error. A confident wrong action is therefore corrected quickly, and the classes of a new task can enter a policy that earlier tasks have made confident. The readout weights start at zero, so the neurons receive no requests in the first step and begin to learn as soon as the readout makes predictions.

From a biophysical perspective, the requests are small currents injected at the active output synapses and at the trigger, and $\Delta \mathbf{v}$ is the change of the membrane potentials once the neuron has reached its new steady state. The driving term $E_i - v_k$ is the driving force of the synapse. Eq.~\eqref{eq:weight} thus combines three quantities that are present at the synapse: presynaptic activity, driving force and the local change of membrane potential. The only assumption that goes beyond the single synapse is that each contact returns requests with a strength equal to its forward weight.

\subsection{Continual Learning}
\label{sec:continual}

The network learns its tasks one after another and does not encounter the data of the earlier tasks again. The learning rule
itself does not use task boundaries: every step is processed in the same way, whichever task it belongs to. Two
mechanisms limit the forgetting of earlier tasks. \emph{Consolidation} slows down synapses that have changed strongly in the recent
past, and the \emph{task code} protects the weights of the branches and neurons that it gates.

Consolidation lowers the learning rate of each synapse according to its own history. With consolidation, the weight
change of Eq.~\eqref{eq:weight} becomes:
\begin{equation}
  \Delta w_i = \frac{\eta}{1 + \kappa\, \Omega_i}\, x_i \left( E_i - v_k \right) \Delta v_k,
  \label{eq:consolidated}
\end{equation}
where $\kappa$ is the consolidation strength (e.g., $10^4$) and $\Omega_i$ is the importance of the synapse. The
importance is a slowly decaying sum of the recent weight changes of the synapse, updated after every step:
\begin{equation}
  \Omega_i \leftarrow (1 - \nu)\, \Omega_i + \left| \Delta w_i \right|.
  \label{eq:importance}
\end{equation}

The decay $\nu$ is small (e.g., 0.0002), so the importance reflects approximately the last $1/\nu$ steps, which is longer than the
training on a single task in the experiments. A synapse that has changed little keeps the full learning rate $\eta$. A
synapse whose recent changes add up to $1/\kappa$ learns at half that rate, and its rate keeps falling as it continues
to change. When a synapse stops changing, its importance decays and its learning rate slowly recovers. The strength
$\kappa$ sets the balance between keeping earlier tasks, which favors a large $\kappa$, and learning new tasks quickly,
which favors a small one. As the synapses used by earlier tasks slow down, later tasks are learned mainly by synapses that have changed little before.

Like elastic weight consolidation~\citep{kirkpatrick2017overcoming} and synaptic
intelligence~\citep{zenke2017continual}, consolidation slows down learning on the weights that earlier tasks have relied
on. It differs from both methods in three respects. The importance measures how much a synapse has recently changed,
without any estimate of how much the loss depends on it. Consolidation acts on the learning rate, instead of adding a penalty on the deviation of the weights from the values they had at the end of earlier tasks. Finally, it needs neither
task boundaries nor stored copies of earlier weights, and every synapse computes it only from its own changes. 

In
biological terms, consolidation resembles metaplasticity, in which the history of a synapse changes its capacity for
later plasticity~\citep{abraham1996metaplasticity}.

Section~\ref{sec:network} introduced the task code as a way to route different tasks through partly different branches
and neurons. The task code also limits how much the weights of a gated branch change. In the second pass, requests for
the global output of a neuron enter at its trigger and reach the segments in proportion to the spread vector
$\boldsymbol{\rho}$ (Eqs.~\ref{eq:db-trigger} and~\ref{eq:shift}). The strong inhibition at the entry of a gated branch
adds a large term to the diagonal of $\mathbf{M}$ at that segment, so little of the spread passes into the branch. Its segments therefore shift little (Figure~\ref{fig:passes}b), its synapses change little by
Eq.~\eqref{eq:consolidated}, and the weights it learned under other tasks are largely kept.

The same protection applies to whole neurons. A neuron that the current task code silences has a global output close to
0, where the slope $z(1 - z)/T$ in Eq.~\eqref{eq:db-trigger} is almost 0, so requests for its global output have almost no effect on it. The protection is partial; requests for local outputs enter directly at their segments
(Eq.~\ref{eq:db-local}) and are not blocked by the gate, so a gated branch whose local outputs are active can still
learn.

The two mechanisms complement each other. The task code gives each task partly distinct branches and neurons, so that
fewer synapses are shared between tasks. Consolidation slows down the synapses that a task has used, including those
that later tasks share. \citet{masse2018alleviating} found the same complementarity in conventional
networks, where context-dependent gating combined with synaptic stabilization was more effective than either method
alone. Several design choices described earlier also limit interference between tasks: the readout has
no bias and keeps a share of random actions (Section~\ref{sec:network}), the thresholds remain fixed after calibration
(Section~\ref{sec:network}), and the readout updates only the prediction of the chosen action
(Section~\ref{sec:learning}).

\subsection{Variants}
\label{sec:variants}

The model of Sections~\ref{sec:neuron}--\ref{sec:continual} is the one used in the main experiments. This subsection
describes the variants used for comparison and in the ablations, and a local learning rule of an earlier version of
the model. They are not part of the model described above.

\paragraph{Gradient-based learning.}
Because the global output is a firing probability, the forward pass is a differentiable function
of all weights and thresholds. The same network can therefore also be trained by backpropagation. Inside a neuron, the
exact gradient follows from two linear systems with the matrix of the forward pass:
\begin{equation}
  \boldsymbol{\lambda} = \mathbf{M}^{-1} \frac{\partial L}{\partial \mathbf{v}}
  \label{eq:adjoint-v}
\end{equation}
\begin{equation}
  \boldsymbol{\mu} = \mathbf{M}^{-1} \frac{\partial L}{\partial \boldsymbol{\rho}}.
  \label{eq:adjoint-rho}
\end{equation}

Here $\partial L / \partial \mathbf{v}$ contains the derivatives of $L$ through the outputs of the neuron, as in
Section~\ref{sec:learning}. The vector $\partial L / \partial \boldsymbol{\rho}$ contains the derivatives through the
back-boost: its entry for an active local output $k$ is $\beta z\, \partial L / \partial y_k$ (Eq.~\ref{eq:boosted}),
and its other entries are 0. The derivative with respect to the weighted input of synapse $i$ on segment $k$ is:
\begin{equation}
  \frac{\partial L}{\partial g_i} = \left( E_i - v_k \right) \lambda_k - \rho_k\, \mu_k.
  \label{eq:exact-gradient}
\end{equation}

The first term has the form that recirculation computes, with $\boldsymbol{\lambda} = -\Delta \mathbf{v}$
(Eqs.~\ref{eq:shift} and~\ref{eq:weight}). The second term describes how the weighted inputs change the spread vector
$\boldsymbol{\rho}$, and with it the back-boost. It is the term that the second pass of recirculation leaves out.

Three objectives are available. The supervised objective is the cross-entropy between the softmax of the values
$Q_a$ and the label, and the network chooses the action with the largest value. It uses the label, so it serves as a
reference with full information. The action-value objective uses the reward only and is the objective of
recirculation learning: the value $Q_a$ of the chosen action is regressed on $r + \gamma V'$, with the target held
fixed. Gradients then reach every synapse by backpropagation instead of the second pass, so the comparison with recirculation keeps the reward signal the same and changes the way in which the weights are updated. The actor-critic objective also uses the reward only. The
outputs $Q_a$ of the readout then serve as the preferences of the softmax policy, and a second linear readout, the
critic, predicts the value $V$ of the current input. The error
$r + \gamma V' - V$ trains the critic and weights the policy gradient of the softmax policy, and a small entropy bonus
keeps the policy from becoming deterministic too early~\citep{sutton2018reinforcement}.

All three objectives use the Adam optimizer~\citep{kingma2015adam} and add a small penalty on the mean global output of every layer, which keeps long-distance activity sparse. After each step, the weights are projected back to the interval between 0 and their upper bound. The thresholds can be learned as well. With the action-value
objective, gradient-based learning therefore differs from recirculation in five respects: the exact gradient including
the second term of Eq.~\eqref{eq:exact-gradient}, the optimizer, the sparsity penalty, the learned thresholds, and the absence of consolidation.

\paragraph{Local learning rule.}
An earlier version of the model learned with a local rule, which recirculation later replaced. In that version, the
trigger produces a sampled spike during training: $z$ is 1 with the probability $p$ given by Eq.~\eqref{eq:global}, and 0 otherwise. The randomness of the spikes provides both the exploration and the signal from which each synapse estimates its contribution to the reward. There is no second pass; instead, a single error is broadcast to every synapse of the
network, the temporal-difference error of the value of the current input:
\begin{equation}
  \bar{\delta} = r + \gamma V' - V,
  \label{eq:broadcast-error}
\end{equation}
where $V = \sum_a \pi_a Q_a$. Unlike the readout error of Eq.~\eqref{eq:error}, $\bar{\delta}$ does not depend on the
action that was taken. Each synapse combines this error with quantities that are available at the synapse, as in
three-factor learning rules~\citep{fremaux2016neuromodulated}. The weight change has two parts, a reward part and a branch
part.

The reward part credits each synapse for its effect on the spike of its neuron:
\begin{equation}
  \Delta w_i^{\mathrm{reward}} = \eta_{\mathrm{r}}\, \bar{\delta}\, \frac{z - p}{T}\, x_i \left( E_i - v_k \right) \rho_k,
  \label{eq:reward-part}
\end{equation}
with learning rate $\eta_{\mathrm{r}}$. The product $x_i (E_i - v_k) \rho_k$ measures how strongly the weight moves the
trigger activation, because $\rho_k$ is the effect of segment $k$ on the trigger (Section~\ref{sec:neuron}). The synapse
has access to $\rho_k$ because the back-boost adds $\beta z \rho_k$ to its segment whenever the neuron fires. The factor
$(z - p)/T$ is positive whenever the neuron fired and negative whenever it stayed silent, and its magnitude is largest when the outcome was unlikely. Together, these factors form the derivative of the log-probability of the
spike with respect to the weight, so for each neuron Eq.~\eqref{eq:reward-part} is the REINFORCE estimate of the policy
gradient~\citep{williams1992simple}, with $\bar{\delta}$ as the reinforcement. In multi-step episodes, the factors after
$\bar{\delta}$ are accumulated in a decaying eligibility trace~\citep{sutton2018reinforcement}.

The reward part credits only what a neuron does through its spike. The branch part trains the local outputs. Each local
output taken before the back-boost, $\hat{y}_k = \max(0, v_k - t_k)$, learns to predict the spike of its own neuron.
The spike reaches the branch through the back-boost, so the target of this prediction is available at the segment. The
prediction error $\delta_k = z + \gamma \hat{y}'_k - \hat{y}_k$ changes the synapses on segment $k$, while the output is
active, by
\begin{equation}
  \Delta w_i^{\mathrm{branch}} = \eta_{\mathrm{b}}\, \delta_k\, x_i \left( E_i - v_k \right),
  \label{eq:branch-part}
\end{equation}
with learning rate $\eta_{\mathrm{b}}$. This part needs no reward. Near the trigger, $\rho_k$ is large and the reward
part dominates; on distal segments, $\rho_k$ is small and the branch part dominates. The two parts are added and
multiplied by the consolidation factor of Eq.~\eqref{eq:consolidated}. The readout learns as in
Eq.~\eqref{eq:nlms}, and the task code protects gated branches as in Section~\ref{sec:continual}: a gated branch
receives little of the back-boost and therefore little reward credit. In addition, each trigger threshold was adapted
during training so that its neuron fired on a target fraction of the steps. This homeostatic mechanism  \citep{leon2026biologically} kept activity sparse and
reactivated silent neurons, but it also changed the representations of earlier tasks.

All credit in this rule comes from the correlation between one broadcast error and the random spike of each neuron. The
random spikes of all other neurons add noise to this correlation, so each synapse receives a weak signal, and the signal
becomes weaker as the network grows. In the experiments, the local rule had little effect on the results.
Recirculation replaces the random spikes by a deterministic second pass, in which every neuron receives its own
requested change.

\paragraph{Options of the model.}
Table~\ref{tab:variants} lists the options that change the default model. Each option changes one element, and the
default model is set when all of them are off.

\begin{table}[t]
  \centering
  \caption{Options of the model used in the experiments. Each row describes the change from the default model.}
  \label{tab:variants}
  \begin{tabular}{lp{0.57\linewidth}}
    \hline
    Option & Change from the default model \\
    \hline
    No task code              & The network receives no task code, so no branch is gated, and all tasks use the same
                                branches and neurons. \\
    Trunk gating              & The task code also inhibits the trigger segment of a random subset of neurons, which
                                silences their global output under that task. \\
    Readout from all layers   & The global outputs of all layers reach the readout, not only those of the last layer
                                (Eq.~\ref{eq:readout}). \\
    Readout bias              & Each action has a bias for the readout, shared by all inputs. \\
    Synapses on branches only & No synapse is placed on the trunk segment. \\
    Outputs at branch tips    & Only the distal segment of each branch has a local output, 4 per neuron instead of
                                8. \\
    No back-boost             & The back-boost strength is $\beta = 0$, so local outputs do not depend on the global
                                output. \\
    No consolidation          & The consolidation strength is $\kappa = 0$, so every synapse keeps the full learning
                                rate. \\
    More synapses per source  & Each sampled source forms several excitatory or inhibitory synapses on the neuron. \\
    Other neuron shapes       & The neuron is an arbitrary tree, with its own segments, leaks, couplings, outputs and
                                gated segments. \\
    \hline
  \end{tabular}
\end{table}

\section{Experimental Results}

\subsection{Setup}
\label{sec:setup}

In these experiments, the same network is trained with different learning methods on two problems, MNIST and Iris, in which tasks are learned one after another. The scenarios compare how well each method handles the current task and how well it remembers the earlier ones.

\paragraph{MNIST.}
The 10 digits are divided into 5 tasks of 2 digits each, learned in the order 0--1, 2--3, 4--5, 6--7 and 8--9.
Each training step presents a batch of 32 images of the current task. Each image is a one-step episode: the network
chooses one of the ten classes and receives a reward of 1 if the class is correct and 0 otherwise. All 10 actions
remain available in every task, so the network can also choose a class of an earlier or a later task. Recirculation is
trained for 500 or 1500 steps per task, and the other learners for 1500 steps, which corresponds to about 4 passes
over the roughly 12,000 training images of a task. The images of earlier tasks do not reappear. After each task, the network is tested
on the test images of every task. At test time, the global outputs produce firing probabilities for every learner, and
the network chooses the action with the largest value.

In the default setting, the network receives the task code presented in Section~\ref{sec:network} during training and testing.
Because the code identifies the task at test time, this setting is task-incremental in the terminology
of~\citet{vandeven2022three}, although the network has a single readout and always chooses among all 10 classes.
Without the task code, the network must recognize the digit without being told which task it belongs to.

\paragraph{Iris.}
The Iris data set~\citep{fisher1936use} contains 150 flowers, 50 of each of 3 species: setosa, versicolor and
virginica. Each flower is described by 4 measurements: the length and width of its sepals and of its petals. Setosa is
linearly separable from the other two species, whereas versicolor and virginica slightly overlap. Each measurement is scaled to $[0, 1]$ with fixed ranges that enclose all flowers, and it is encoded by 8 Gaussian tuning curves with evenly spaced centers \citep{leon2026sparse} and a width equal to their spacing, so the network receives 32 inputs. Flowers of different sizes therefore excite largely different inputs. 15 instances of each class are held out for testing, with a new split in every run.

The sequence has 3 tasks: setosa and versicolor, then versicolor and virginica, and finally all three species.
Setosa is absent from the second task, so the accuracy on the first task after the second one measures forgetting. All
3 actions are available in every task, and no task code is used. The network has a single layer of 64 neurons.
The inputs have no positions, so each neuron samples 12 of the 32 inputs from all of them, and the readout reads the global outputs of the layer. Training lasts 300 steps per task, with batches of 16 flowers.

\paragraph{Learning methods.}
Four learning methods are compared on the same network, built and calibrated in the same way:
\begin{itemize}[noitemsep]
  \item \emph{Recirculation}, the model of Sections~\ref{sec:learning} and~\ref{sec:continual}, with learning rate  $\eta = 20$ in Eq.~\eqref{eq:consolidated}, consolidation strength $\kappa = 10^4$ and decay $\nu = 0.0002$;
  \item \emph{Gradient-based learning} (Section~\ref{sec:variants}), with the supervised and the action-value
        objectives. Adam updates all weights, including the readout, with learning rate $10^{-3}$, $\beta_1 = 0.9$ and     $\beta_2 = 0.999$. The weight of the sparsity penalty is $10^{-3}$, the thresholds are learned, and there is no  consolidation;
  \item The \emph{local rule} (Section~\ref{sec:variants}), with learning rates $\eta_r = 0.002$ and
        $\eta_b = 0.0005$ and consolidation strength $\kappa = 10^3$. The adaptation of the trigger
        thresholds is switched off, because it shifted the representations of earlier tasks;
  \item A \emph{fixed network}: the local rule with both learning rates set to zero. The neurons keep their calibrated weights, and only the readout learns. This learner shows how much the structure of the network contributes by itself.
\end{itemize}

\paragraph{Performance measures.}
Let $R_{t,k}$ be the test accuracy on task $k$ after training on task $t$, for $t, k = 1, \dots, N$, where $N$ is the
number of tasks. The results are summarized by four measures:

\begin{itemize}[noitemsep]
  \item The \emph{accuracy after training}, $\frac{1}{N}\sum_{k} R_{k,k}$, the mean accuracy on each task right after it is learned. It measures how well a learning method acquires new tasks;
  \item The \emph{final accuracy}, $\frac{1}{N}\sum_{k} R_{N,k}$, the mean accuracy on all tasks at the end of
        training;
  \item The \emph{forgetting}, $\frac{1}{N-1}\sum_{k<N} \bigl( \max_{k \le t < N} R_{t,k} - R_{N,k} \bigr)$, the mean loss of accuracy of each earlier task between its best value and the end of training~\citep{chaudhry2018riemannian};
  \item The \emph{early training reward}, the mean reward over the first 300 training steps of each task (25 steps
        for Iris), averaged over the tasks. It includes exploration and measures how quickly new tasks are learned.
\end{itemize}

On Iris, the tasks share classes, so the averages over tasks count some classes more than once. The Iris results are
therefore reported as individual entries of $R_{t,k}$.

The values in the tables are means over 10 runs of each configuration, and the figures show example runs.

\subsection{Gradient-Based Learning and Recirculation}
\label{sec:gradient-vs-recirculation}

\begin{figure}[t]
  \centering
  \includegraphics[width=\linewidth]{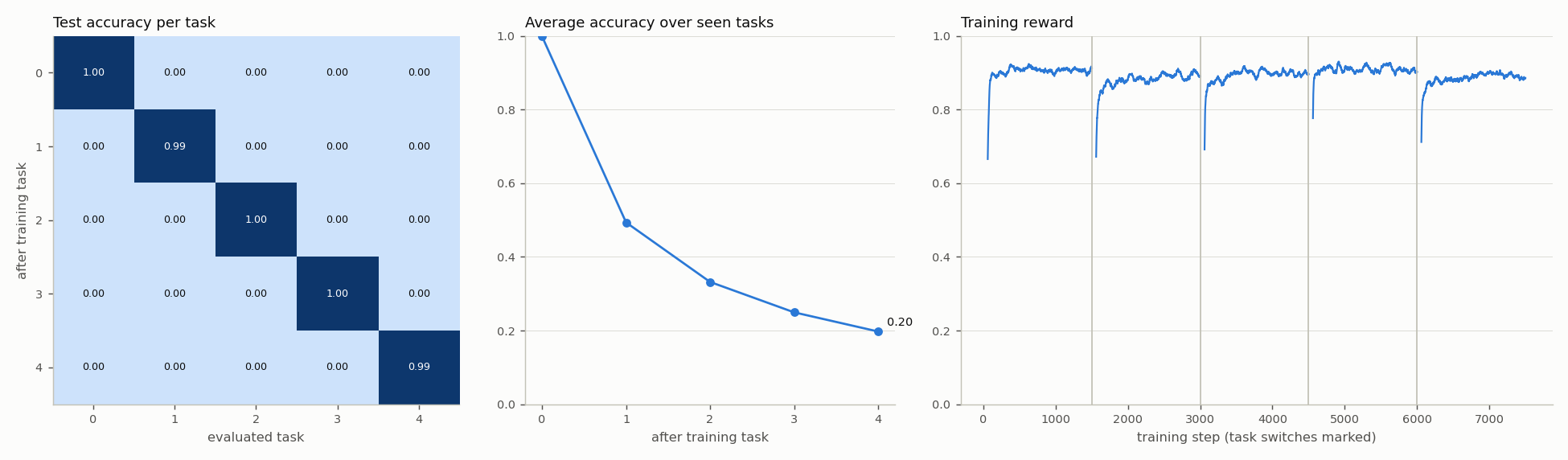}\\[2pt]
  {\small (a) Gradient-based learning, action-value objective, 1500 steps per task}\\[8pt]
  \includegraphics[width=\linewidth]{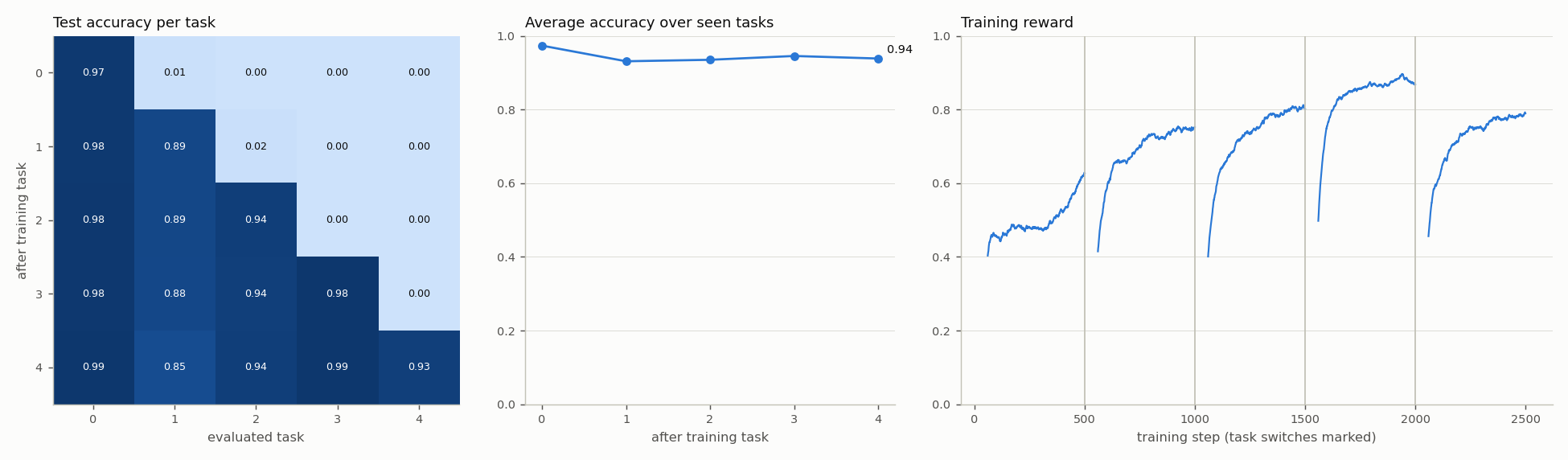}\\[2pt]
  {\small (b) Recirculation, 500 steps per task}
  \caption{Example runs on MNIST with the task code. Left: test accuracy on each task (columns) after training on each
    task (rows); the tasks are numbered from 0 in the order of training. Middle: mean accuracy over the tasks seen so
    far. Right: training reward, with the task boundaries marked.}
  \label{fig:main}
\end{figure}

Table~\ref{tab:main} compares gradient-based learning and recirculation on MNIST with the task code, and
Figure~\ref{fig:main} shows an example run of each.

\begin{table}[t]
  \centering
  \caption{Gradient-based learning and recirculation on MNIST with the task code. After training: mean accuracy on
    each task right after it is learned. Final: mean accuracy on all tasks at the end of training. Forgetting: mean
    loss of accuracy of the earlier tasks. Early reward: mean training reward in the first 300 steps of each task. Steps: training steps per task. With
    the supervised objective, the network does not explore, so its training reward is not comparable with that of the
    reward-based learners.}
  \label{tab:main}
  \begin{tabular}{lccccc}
    \hline
    Learner & Steps & After    & Final & Forgetting & Early  \\
            &       & training &       &            & reward \\
    \hline
    Gradient, supervised    & 1500 & 0.993 & 0.197 & 0.994 & --    \\
    Gradient, action values & 1500 & 0.994 & 0.198 & 0.995 & 0.846 \\
    Recirculation           &  500 & 0.944 & 0.939 & 0.009 & 0.632 \\
    Recirculation           & 1500 & 0.966 & 0.868 & 0.123 & 0.611 \\
    \hline
  \end{tabular}
\end{table}

Gradient-based learning acquires every task almost perfectly and quickly. With either objective, its accuracy after
training is 0.99. With the action-value objective, the training reward rises to about 0.9 within roughly a hundred
steps of each new task (Figure~\ref{fig:main}a, right), close to the limit of 0.91 set by the share
$\varepsilon = 0.1$ of random actions. However, each new task erases all earlier ones. After each task, the accuracy on
all earlier tasks drops to zero (Figure~\ref{fig:main}a, left), and the mean accuracy over the tasks seen so far falls
as $1/t$ (middle). The final accuracy, 0.197 with the supervised objective and 0.198 with the action-value objective, is
the accuracy on the last task divided by five. This occurs both with the label and with the reward alone, even though the task code gates a different subset of branches in each task.

Recirculation retains the tasks it has learned. With 500 steps per task, its forgetting is 0.009, more than a hundred times lower
than that of gradient-based learning, and its final accuracy of 0.939 is almost equal to its accuracy right after
training, 0.944. All earlier tasks remain close to their initial accuracy until the end of the sequence
(Figure~\ref{fig:main}b, left), and the mean accuracy over the tasks seen so far stays flat (middle). Recirculation
achieves this only from the reward, with quantities available at each synapse, without task boundaries in the learning rule, and without stored copies of earlier weights.

Longer training raises the accuracy of recirculation on each task. With 1500 steps per task, the accuracy after
training is 0.966, close to the 0.99 of gradient-based learning, while the forgetting, 0.123, remains about eight times
lower than that of gradient-based learning. The final accuracy is 0.868, more than four times the 0.198 of
gradient-based learning with the same reward signal. The length of training and the strength of consolidation set the
balance between learning new tasks and keeping earlier ones (Section~\ref{sec:tradeoff}).

The action-value objective uses the same reward signal as recirculation, and with it, gradient-based learning reaches the same accuracy and the same forgetting as with the label. The difference in forgetting between the two learning methods therefore comes from the way in which recirculation changes the weights, and not from the feedback received.

\subsection{Stability--Plasticity Trade-off}
\label{sec:tradeoff}

\begin{figure}[t]
  \centering
  \includegraphics[width=0.75\linewidth]{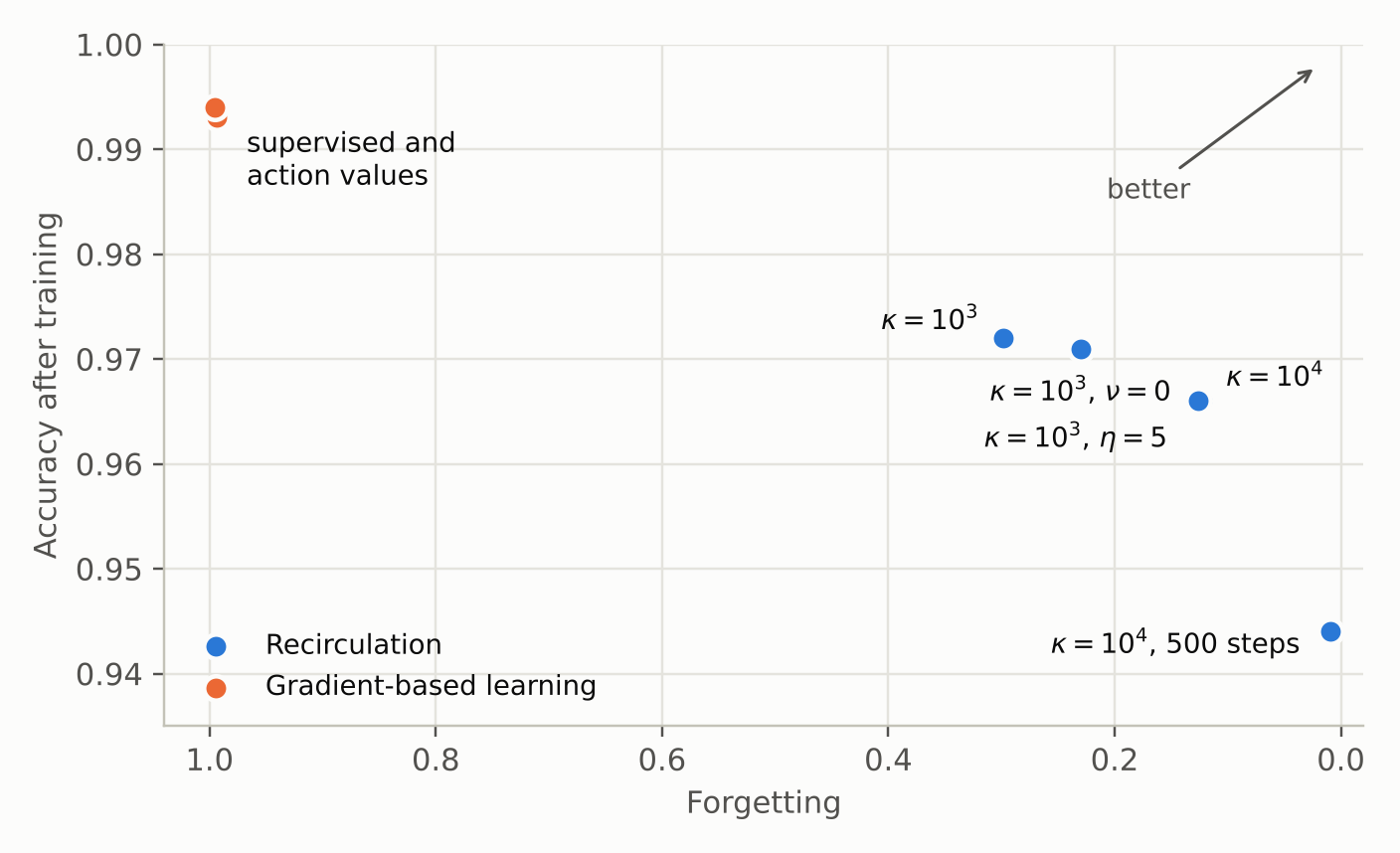}
  \caption{Accuracy after training and forgetting on MNIST with the task code, for recirculation with the settings of
    Table~\ref{tab:tradeoff} and for gradient-based learning with both objectives. Unless marked otherwise,
    $\nu = 0.0002$, $\eta = 20$ and training lasts 1500 steps per task. The settings $\kappa = 10^4$ and
    $\kappa = 10^3$ with $\eta = 5$ give almost the same result.}
  \label{fig:tradeoff}
\end{figure}

Three parameters of recirculation determine how far a synapse moves while a task is learned and how much of this
change later tasks can undo: the learning rate $\eta$, the consolidation strength $\kappa$ and the decay $\nu$ of the importance (Eqs.~\ref{eq:consolidated} and~\ref{eq:importance}). The number of training steps per task acts in a similar way. Table~\ref{tab:tradeoff} and Figure~\ref{fig:tradeoff} show how these settings place recirculation between learning each new task as well as possible and keeping all earlier tasks.

\begin{table}[t]
  \centering
  \caption{Recirculation on MNIST with the task code, for different consolidation strengths $\kappa$, importance decays
    $\nu$, learning rates $\eta$ and numbers of training steps per task. Gradient-based learning with the action-value
    objective is shown for comparison. The measures are those of Table~\ref{tab:main}.}
  \label{tab:tradeoff}
  \begin{tabular}{ccccccc}
    \hline
    $\kappa$ & $\nu$ & $\eta$ & Steps & After    & Final & Forgetting \\
             &       &        &       & training &       &            \\
    \hline
    $10^4$ & 0.0002 & 20 &  500 & 0.944 & 0.939 & 0.009 \\
    $10^4$ & 0.0002 & 20 & 1500 & 0.966 & 0.868 & 0.123 \\
    $10^3$ & 0.0002 &  5 & 1500 & 0.966 & 0.866 & 0.126 \\
    $10^3$ & 0      & 20 & 1500 & 0.971 & 0.786 & 0.230 \\
    $10^3$ & 0.0002 & 20 & 1500 & 0.972 & 0.733 & 0.298 \\
    \hline
    \multicolumn{3}{l}{Gradient, action values} & 1500 & 0.994 & 0.198 & 0.995 \\
    \hline
  \end{tabular}
\end{table}

Stronger consolidation reduces forgetting at almost no cost in accuracy on the current task. With 1500 steps per task,
raising $\kappa$ from $10^3$ to $10^4$ lowers the forgetting from 0.298 to 0.123 and raises the final accuracy from
0.733 to 0.868, while the accuracy after training changes only from 0.972 to 0.966. Consolidation slows down mainly the
synapses that have changed strongly in the recent past, so the synapses that a new task needs still learn at almost
the full rate.

The other parameters have similar effects. An importance that never decays ($\nu = 0$) lowers the forgetting at
$\kappa = 10^3$ from 0.298 to 0.230. A learning rate of $\eta = 5$ instead of 20 has almost the same effect as the
stronger consolidation, with a forgetting of 0.126 and an accuracy after training of 0.966. Shorter training gives the
best final accuracy: with $\kappa = 10^4$ and 500 steps per task, the forgetting is 0.009 and the final accuracy 0.939.

Across all settings, the accuracy after training stays between 0.944 and 0.972, within 0.05 of gradient-based
learning, while the forgetting ranges from 0.009 to 0.298, less than a third of the 0.995 of gradient-based learning.
The final accuracy ranges from 0.733 to 0.939, compared with 0.198. Recirculation thus offers a continuous control of
the balance between learning new tasks and keeping earlier ones, through parameters that each synapse applies to its
own changes. Gradient-based learning without consolidation lies at one extreme of this trade-off, where each new task is
learned best and all earlier tasks are lost (Figure~\ref{fig:tradeoff}, upper left).

\subsection{Ablations}
\label{sec:ablations}

\begin{figure}[t]
  \centering
  \includegraphics[width=\linewidth]{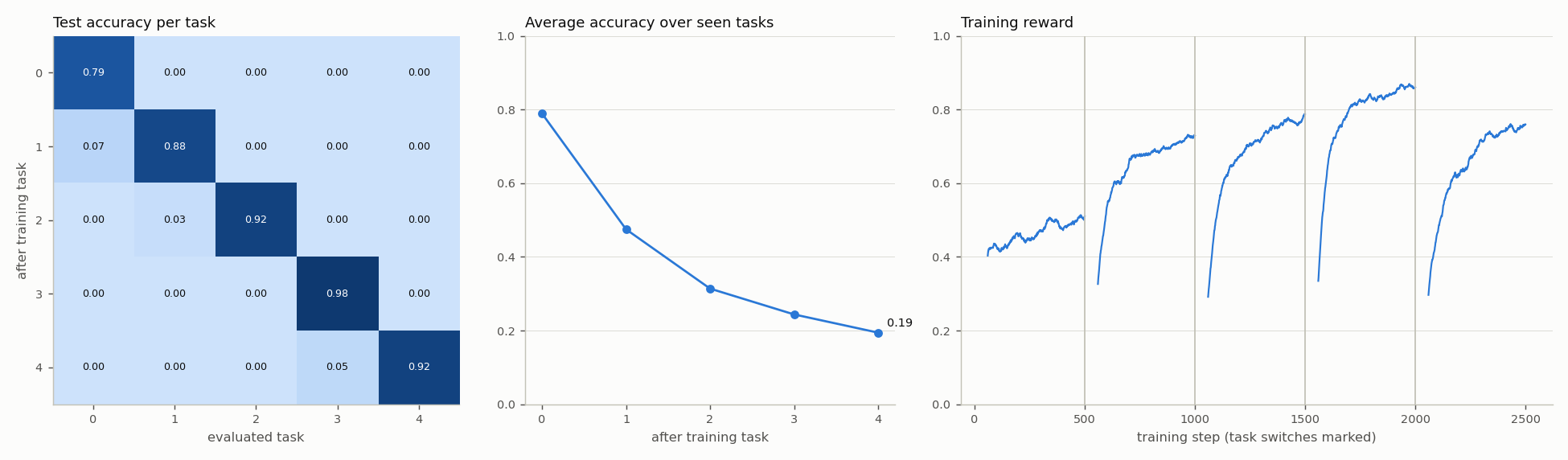}\\[2pt]
  {\small (a) Recirculation without the task code, 500 steps per task}\\[8pt]
  \includegraphics[width=\linewidth]{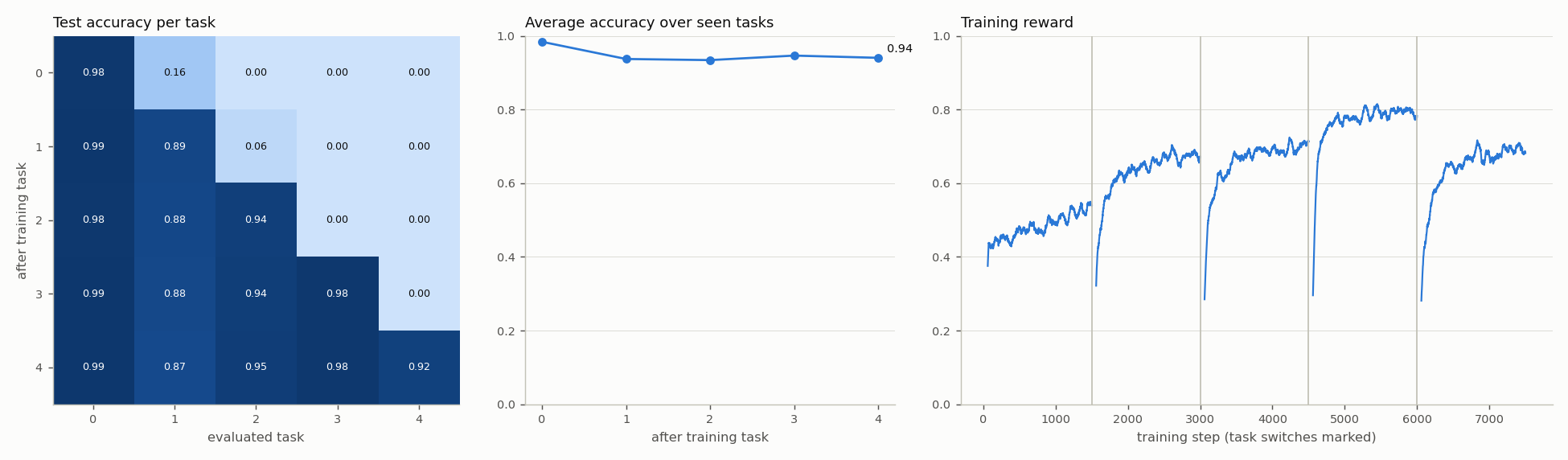}\\[2pt]
  {\small (b) Local rule with the task code, 1500 steps per task}
  \caption{Example runs of ablations on MNIST. The panels are those of Figure~\ref{fig:main}.}
  \label{fig:ablations}
\end{figure}

The ablations presented in this section change one element of the model at a time. Table~\ref{tab:ablations-recirculation} lists the ablations of
recirculation and Table~\ref{tab:ablations-local} those of the local rule, and Figure~\ref{fig:ablations} shows an
example run of each learning method.

\begin{table}[t]
  \centering
  \caption{Ablations of recirculation on MNIST. Each group compares the default model with changes of one element, at
    the consolidation strength $\kappa$ and the number of training steps per task given in the row. No task code: the
    network receives no task code in training and testing. Readout from all layers: the global outputs of both layers
    reach the readout. Single layer: the network has only the first layer, whose global outputs reach the readout.
    80\% of branches gated: each code unit gates the entry segment of a branch with probability 0.8 instead of 0.5.
    The measures are those of Table~\ref{tab:main}.}
  \label{tab:ablations-recirculation}
  \begin{tabular}{lccccc}
    \hline
    Configuration & $\kappa$ & Steps & After    & Final & Forgetting \\
                  &          &       & training &       &            \\
    \hline
    Default                  & $10^4$ &  500 & 0.944 & 0.939 & 0.009 \\
    No task code             & $10^4$ &  500 & 0.895 & 0.195 & 0.875 \\
    Readout from all layers  & $10^4$ &  500 & 0.969 & 0.297 & 0.840 \\
    \hline
    Default                  & $10^4$ & 1500 & 0.966 & 0.868 & 0.123 \\
    Single layer             & $10^4$ & 1500 & 0.980 & 0.203 & 0.971 \\
    \hline
    Default                  & $10^3$ & 1500 & 0.972 & 0.733 & 0.298 \\
    80\% of branches gated   & $10^3$ & 1500 & 0.972 & 0.602 & 0.463 \\
    \hline
  \end{tabular}
\end{table}

\begin{table}[t]
  \centering
  \caption{Ablations of the local rule on MNIST with the task code, with 1500 steps per task and consolidation
    strength $\kappa = 10^3$. No reward part: $\eta_{\mathrm{r}} = 0$ in Eq.~\eqref{eq:reward-part}. No branch part:
    $\eta_{\mathrm{b}} = 0$ in Eq.~\eqref{eq:branch-part}. Readout only: both parts are removed, which gives the fixed
    network of Section~\ref{sec:setup}. No consolidation: $\kappa = 0$. The measures are those of
    Table~\ref{tab:main}.}
  \label{tab:ablations-local}
  \begin{tabular}{lcccc}
    \hline
    Configuration & After    & Final & Forgetting & Early  \\
                  & training &       &            & reward \\
    \hline
    Full rule        & 0.939 & 0.934 & 0.008 & 0.520 \\
    No reward part   & 0.939 & 0.933 & 0.009 & 0.521 \\
    No branch part   & 0.936 & 0.931 & 0.008 & 0.510 \\
    Readout only     & 0.936 & 0.931 & 0.008 & 0.510 \\
    No consolidation & 0.947 & 0.326 & 0.777 & 0.572 \\
    No task code     & 0.934 & 0.223 & 0.889 & 0.421 \\
    \hline
  \end{tabular}
\end{table}

The task code is essential for remembering earlier tasks. Without it, recirculation still learns each task, with an
accuracy after training of 0.895, but it forgets 0.875 of it, and the final accuracy falls to 0.195. The accuracy
matrix then becomes almost diagonal (Figure~\ref{fig:ablations}a), as that of gradient-based learning
(Figure~\ref{fig:main}a). The local rule behaves in the same way, with a forgetting of 0.889 without the code. With the
code, each task drives its own subset of branches and neurons, and the gating limits how much the weights of the other
subsets change (Section~\ref{sec:continual}). 
Gradient-based learning forgets earlier tasks even with the task code (Section~\ref{sec:gradient-vs-recirculation}), and so does the local rule without consolidation (forgetting 0.777, Table 7), so it is the combination of the task code and consolidation that helps remember earlier tasks.

The readout usually receives only the global outputs of the last layer. When the outputs of the first layer also reach the readout,
recirculation acquires each task better, with an accuracy after training of 0.969, but the forgetting rises to 0.840.
A network with a single layer forgets 0.971. A likely reason is that the global outputs of the second layer depend on
the task code twice, through the gating of their own branches and through the gated outputs of the first layer that
they receive, so that they are more specific to each task than those of the first layer.

Gating half of the branches is more effective than gating most of them. With 80\% of the branches gated, the accuracy after
training is unchanged, but the forgetting rises from 0.298 to 0.463. The default fraction of 0.5 is therefore used in
all other experiments.

With the local rule, the structure of the network protects the earlier tasks by itself. Even when only the readout
learns, the forgetting is 0.008, and the final accuracy is 0.931 (Table~\ref{tab:ablations-local}). The two parts of
the local rule change these results little: removing the reward part, the branch part or both changes every accuracy by
at most 0.003. The credit signal of the local rule is weak, because each synapse correlates one broadcast error with
the random spike of its own neuron, and the random spikes of all other neurons add noise to this correlation
(Section~\ref{sec:variants}). Consolidation is what retains the earlier tasks when the rule does change the weights:
without it, the accuracy after training rises to 0.947, but the forgetting rises to 0.777.

Recirculation keeps the protection given by the structure of the network while it trains the neurons. With 500 steps
per task, its forgetting is as low as that of the fixed network (0.009 compared with 0.008), and its final accuracy is
higher (0.939 compared with 0.931). With 1500 steps per task, its accuracy after training rises to 0.966, compared with
0.936 for the fixed network. In recirculation, every neuron receives its own requested change through the second pass,
instead of a single broadcast error, so the training of the neurons increases the accuracy on each task.

\subsection{Iris}
\label{sec:iris}

\begin{figure}[t]
  \centering
  \includegraphics[width=\linewidth]{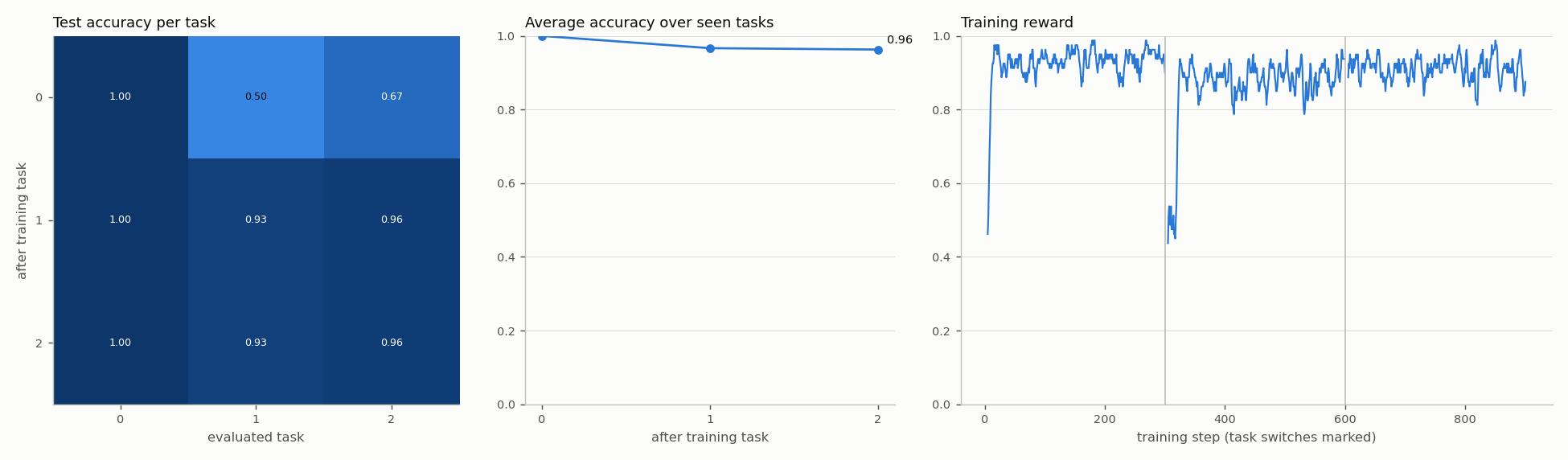}
  \caption{Example run of recirculation on Iris. Task 0: setosa and versicolor. Task 1: versicolor and virginica.
    Task 2: all three species. The panels are those of Figure~\ref{fig:main}.}
  \label{fig:iris}
\end{figure}

On Iris, the network has a single layer of 64 neurons and receives no task code. Table~\ref{tab:iris} gives the test
accuracy on each task after training on each task, and Figure~\ref{fig:iris} shows an example run.

\begin{table}[t]
  \centering
  \caption{Test accuracy $R_{t,k}$ on Iris. Rows: after training on task $t$. Columns: tested task $k$. The entries
    above the diagonal are measured before task $k$ is trained.}
  \label{tab:iris}
  \begin{tabular}{lccc}
    \hline
    After training on     & Setosa,    & Versicolor, & All three \\
                          & versicolor & virginica   & species   \\
    \hline
    Setosa, versicolor    & 1.000      & 0.500       & 0.667     \\
    Versicolor, virginica & 0.980      & 0.947       & 0.964     \\
    All three species     & 0.980      & 0.947       & 0.964     \\
    \hline
  \end{tabular}
\end{table}

The network learns the first task, setosa and versicolor, perfectly. The training reward rises to about 0.9 within
roughly 25 steps (Figure~\ref{fig:iris}, right), close to the limit set by the random actions. Before the second task
is trained, the accuracy on it is already 0.500, because the versicolor flowers of the second task are recognized from
the first task.

The second task, versicolor and virginica, contains the two species that overlap. The network learns it to an accuracy
of 0.947, and it keeps the first task, although no setosa flower is shown during the second task and no task code
indicates which task is being trained: the accuracy on setosa and versicolor decreases only from 1.000 to 0.980. When the second task
starts, the training reward drops to about 0.5, because virginica flowers are first classified as versicolor, and it
recovers within a few tens of steps.

After the second task, the network already classifies all three species with an accuracy of 0.964, although it has
never seen setosa and virginica together. The two tasks, learned one after the other, combine into a three-way
classifier without any joint training. The third task, which shows all three species together, leaves every test
accuracy unchanged, and its training reward stays high from its first step.

Recirculation also retains earlier tasks when they are presented in a different order. When the network learns versicolor and
virginica first, then setosa and versicolor, and finally setosa and virginica, the accuracy on the last task is already
0.900 before it is trained, the final accuracy is 0.933, and the forgetting is 0.083.

\subsection{Discussion}
\label{sec:discussion}

The results point to three features of the model:

\begin{itemize}
\item A large part of the protection of earlier tasks follows from the structure of the neurons rather than from an added mechanism;
\item The neurons can be trained with signals available at each synapse without losing this protection;
\item The balance between learning new tasks and keeping earlier ones is set by parameters that each synapse applies to its own changes.
\end{itemize}

In conventional networks, protection against forgetting is usually added to the learning algorithm, as a penalty on
weight changes, a mask on the gradients or a buffer of stored examples. In the proposed model, much of it is a
consequence of the neuron model. The task code acts through shunting inhibition at the entry of a branch, and because
inhibition is divisive and acts where it is placed (Eq.~\ref{eq:divisive}), a gated branch is separated from the trunk
in both passes. In the first pass, it contributes little to the global output. In the second pass, it receives little
of the requested change, because a strongly inhibited segment has small entries in the spread vector $\boldsymbol{\rho}$.
The same inhibition that routes a task through a subset of branches thus shields the other subsets from learning, and no mask has to be stored or applied to the weight changes. The readout adds to this protection, because it has no bias and changes only the value of the chosen action, in proportion to the global outputs that are active. With these elements alone, a network whose neurons do not learn keeps its earlier tasks almost completely, with a forgetting below 0.01.

The same structure does not protect earlier tasks under gradient-based learning, whose forgetting is almost complete even with the task code. Inside each neuron, the exact gradient is attenuated in gated branches in the same way as the second pass of recirculation, because both are computed with the same matrix. 
Some properties of the gradient-based learner work against this attenuation. Adam normalizes the update of each weight by the size of its recent gradients,
so small gradients in gated branches still produce steps of normal size. The thresholds are learned and act as biases,
which shift with every task. Finally, nothing slows down the weights on which earlier tasks rely. In recirculation, the
update of each weight is proportional to the shift of its own segment, so gated branches change little, the thresholds
remain fixed, and consolidation slows down the weights that have already changed. Before consolidation, recirculation
follows the direction of the gradient, except for the effect of the weights on the back-boost, but it keeps the
attenuation that the structure imposes. The neurons can therefore be trained, which raises the accuracy on each task
above that of the fixed network, while most of the protection is kept. The local rule marks the other extreme: its
broadcast error gives each synapse too weak a signal to improve on the fixed network.

All quantities used by recirculation are available at the synapse: the presynaptic activity, the driving force and
the change of the activation of its own segment between the two passes. Because the matrix of each neuron is
symmetric, requested changes spread through the tree by the same equations as activity, and no separate backward
computation is needed inside a neuron. Consolidation is computed by each synapse from its own history. It needs no task
boundaries, no stored copies of earlier weights and no estimate of how much the loss depends on each weight. Because it
acts on the learning rate and not on the weights, it never freezes a synapse: a synapse that stops changing slowly
regains its plasticity as its importance decays. The cost of a neuron is dominated by its synapses, as for a point
neuron, and the exact solution of its tree adds a term linear in the number of segments; the second pass costs about
as much as the first.

The model combines in a single neuron mechanisms that are usually treated separately. The task code is a form of
context-dependent gating~\citep{masse2018alleviating}, applied to the branches within each neuron by dendritic
inhibition instead of to whole units. Consolidation follows the idea of elastic weight
consolidation~\citep{kirkpatrick2017overcoming} and synaptic intelligence~\citep{zenke2017continual}, but as a
metaplastic change of the learning rate~\citep{abraham1996metaplasticity} that each synapse computes from its own changes. The learning rule belongs to the family of 
recirculation~\citep{hinton1988learning,oreilly1996biologically}
and equilibrium propagation~\citep{scellier2017equilibrium}, in which weights change according to the difference
between two states of the network. Here, the two states are steady states of passive dendritic trees, and the network learns from reward instead of from target outputs. The masking of gated branches and the locality of the updates, which other approaches add as separate components, follow from the biophysics of the neuron.

In terms of limitations, on MNIST the protection of earlier tasks relies on the task code. Without it, the network forgets almost as much as with gradient-based learning (0.875 compared with 0.995), so on such data the model in its present form addresses the task-incremental setting. Deriving the code
from the input, or from the global outputs of the network itself, would extend the protection to the class-incremental setting. The model also trades some plasticity for retention: each task is learned more slowly and to a slightly lower
accuracy than with gradient-based learning. The protection also requires that the readout read only the last layer,
because the global outputs of the first layer are not protected as well. The thresholds are fixed after calibration on
images of the first task, so later tasks are represented with thresholds that do not reflect their statistics. Finally,
each contact returns requested changes with a strength equal to its forward weight. This is the only assumption of the learning rule that goes beyond the single synapse, and it plays the role that weight symmetry plays in backpropagation. Natural extensions of the model are a task code generated by the network, tasks with multi-step episodes, for which the learning rule is already defined, and deeper networks in which every layer contributes to the protection.

\section{Conclusions}

This paper introduced a network of unipolar neurons that learns a sequence of tasks only from reward. Each neuron is modeled as a tree of passive segments with graded local outputs and one global output. The network learns with recirculation, a two-pass rule in which each synapse uses only quantities available at its own location. A task code inhibits a different subset of branches in each task, and consolidation slows down the synapses that have changed strongly in the recent past.

On the sequence of MNIST tasks with the task code, recirculation learned each new task to an accuracy close to that of gradient-based training and remembered the earlier tasks very well. Gradient-based training of the same network forgot the earlier tasks almost completely. The consolidation strength, the learning rate and the length of training determine the balance between learning new tasks and retaining earlier ones. The ablations showed that the task code is necessary for retention and that, in general, the readout should receive only the global outputs of the last layer. On Iris, the network learned tasks with overlapping classes in sequence without a task code, and the tasks combined into a classifier of all three species without joint training.

The most important extension is to infer the task code instead of providing it. This would make the model applicable to the class-incremental setting, in which the identity of the task is unknown at test time. One possibility is a separate group of neurons whose global outputs identify the context of the input. Another is to assign a new code when the error of the reward prediction increases sharply, since such an increase can indicate a change of task.

A second direction is learning in tasks with several steps and delayed reward. The learning rule includes a discount factor, but all experiments used single-step episodes. The control of a simulated flexible arm, similar to the arms of an octopus, could be a suitable test case. Local outputs could coordinate neighboring parts of the arm, and global outputs could convey commands across the whole network.

Several fixed parts of the model could also be learned. Currently, the locations of the synapses and the shapes of the trees are random. Structural plasticity, i.e., the formation and removal of synapses \citep{leon2026biologically}, could place inputs on the branches where they contribute most to the task.

The requested changes could be returned through separate feedback weights instead of the forward weights. This would show how much of the accuracy and of the protection depends on the symmetry between the two directions. A theoretical analysis could also relate the interference between tasks to the strength of the inhibition from the task code and to the fraction of branches it reaches.

Finally, the model should be evaluated with deeper networks, longer task sequences and more complex data.

%
%
%
%
%
%
%
%

\vskip 0.2in
\bibliography{sample}

\end{document}